\documentclass[acmtog, screen, nonacm]{acmart}
\usepackage{threeparttable}
\usepackage[table,dvipsnames]{xcolor}
\usepackage{float}
\usepackage{booktabs}
\usepackage{multirow}
\usepackage{caption}
\usepackage{subcaption}
\usepackage[percent]{overpic}

\usepackage{amsmath}      
\usepackage{bm}
\usepackage{stmaryrd}
\usepackage{bbm}
\usepackage{nicefrac}
\usepackage{empheq}

\usepackage{algorithm}
\usepackage{algorithmicx}
\usepackage{algpseudocode}

\usepackage{acronym}
\usepackage[normalem]{ulem}
\usepackage{wrapfig}
\usepackage{listings}

\usepackage{cleveref}

\definecolor{codegreen}{rgb}{0,0.6,0}
\definecolor{codegray}{rgb}{0.5,0.5,0.5}
\definecolor{codepurple}{rgb}{0.58,0,0.82}
\definecolor{backcolour}{rgb}{0.96,0.96,0.98}

\setcopyright{none}

\newcommand{\apcell}[1]{\cellcolor{Apricot}{#1}}
\newcommand{\tecell}[1]{\cellcolor{teal!25}{#1}}

\makeatletter
\let\@authorsaddresses\@empty
\makeatother

\begin{document}

\title{Birth of a Painting: Differentiable Brushstroke Reconstruction}

\author{Ying Jiang$^{1*}$}
\author{Jiayin Lu$^{1*}$}
\author{Yunuo Chen$^{1*}$}
\author{Yumeng He$^{1,2}$}
\author{Kui Wu$^{3}$}
\author{Yin Yang$^{4}$}
\author{Chenfanfu Jiang$^{1}$}

\renewcommand{\shortauthors}{Ying Jiang, et al.}
\newcommand\blfootnote[1]{%
  \begingroup
  \renewcommand\thefootnote{}\footnote{#1}%
  \addtocounter{footnote}{-1}%
  \endgroup
}

\begin{abstract}

\blfootnote{* indicates equal contributions. \\Affiliations: $^1$UCLA, $^2$USC, $^3$LIGHTSPEED, $^4$Utah}

Painting embodies a unique form of visual storytelling, where the creation process is as significant as the final artwork. Although recent advances in generative models have enabled visually compelling painting synthesis, most existing methods focus solely on final image generation or patch-based process simulation, lacking explicit stroke structure and failing to produce smooth, realistic shading. In this work, we present a differentiable stroke reconstruction framework that unifies painting, stylized texturing, and smudging to faithfully reproduce the human painting–smudging loop. Given an input image, our framework first optimizes single- and dual-color Bézier strokes through a parallel differentiable paint renderer, followed by a style generation module that synthesizes geometry-conditioned textures across diverse painting styles. We further introduce a differentiable smudge operator to enable natural color blending and shading. Coupled with a coarse-to-fine optimization strategy, our method jointly optimizes stroke geometry, color, and texture under geometric and semantic guidance. Extensive experiments on oil, watercolor, ink, and digital paintings demonstrate that our approach produces realistic and expressive stroke reconstructions, smooth tonal transitions, and richly stylized appearances, offering a unified model for expressive digital painting creation. 
\end{abstract}

\begin{teaserfigure}
  \includegraphics[width=\textwidth]{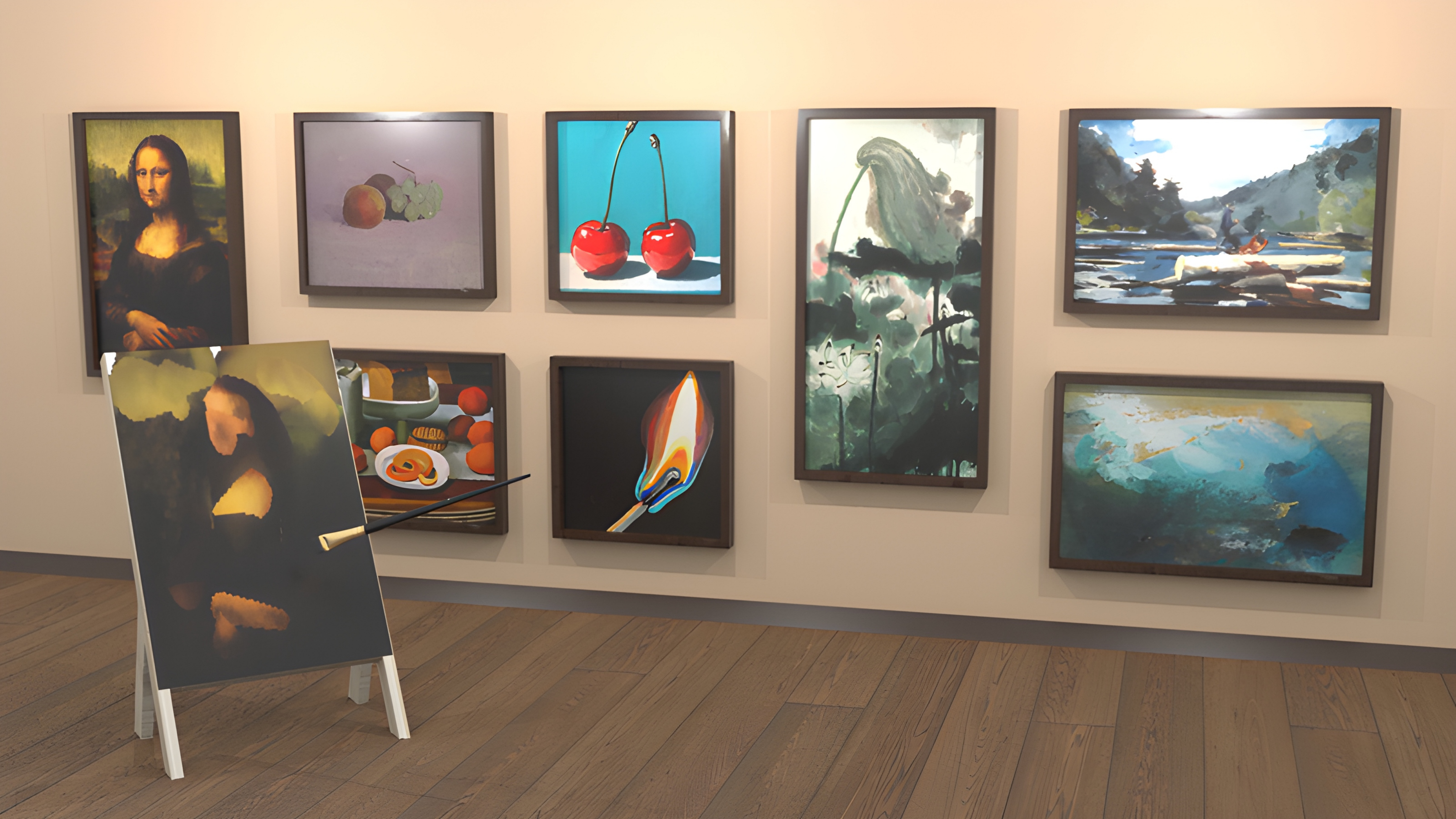}
  \vspace{-0.2in}
  \caption{\textbf{Painting Gallery.} We visualize the reconstructed paintings produced by our method, along with one example of an intermediate painting stage. Our method reconstructs the painting process by progressively generating realistic brushstrokes, resulting in visually coherent and complete paintings.}
  \label{fig:teaser}
\end{teaserfigure}

\maketitle

\section{Introduction}
\label{sec:intro}

As Picasso once said, “Painting is just another way of keeping a diary.” Painting serves as a powerful medium for expressing human emotions and ideas in visual form. Most prior work on painting generation focuses on the finished artwork \cite{galerne2024scaling, ren2024opt, tian2022text}, while the creative process itself remains concealed. Yet, observing how painters construct their work is equally meaningful. It not only allows audiences to appreciate and study the historical context and expressive techniques of art but also serves as a rich resource for learning how to paint for both humans and robots in reality \cite{chen2025spline, schaldenbrand2024cofrida}. Since mastering painting requires substantial time and effort, and much instruction still depends on in-person teaching, generating painting processes offers new and more accessible pathways for learning and engaging with art.

\begin{figure}[t]
\centering  \includegraphics[width=0.45\textwidth]{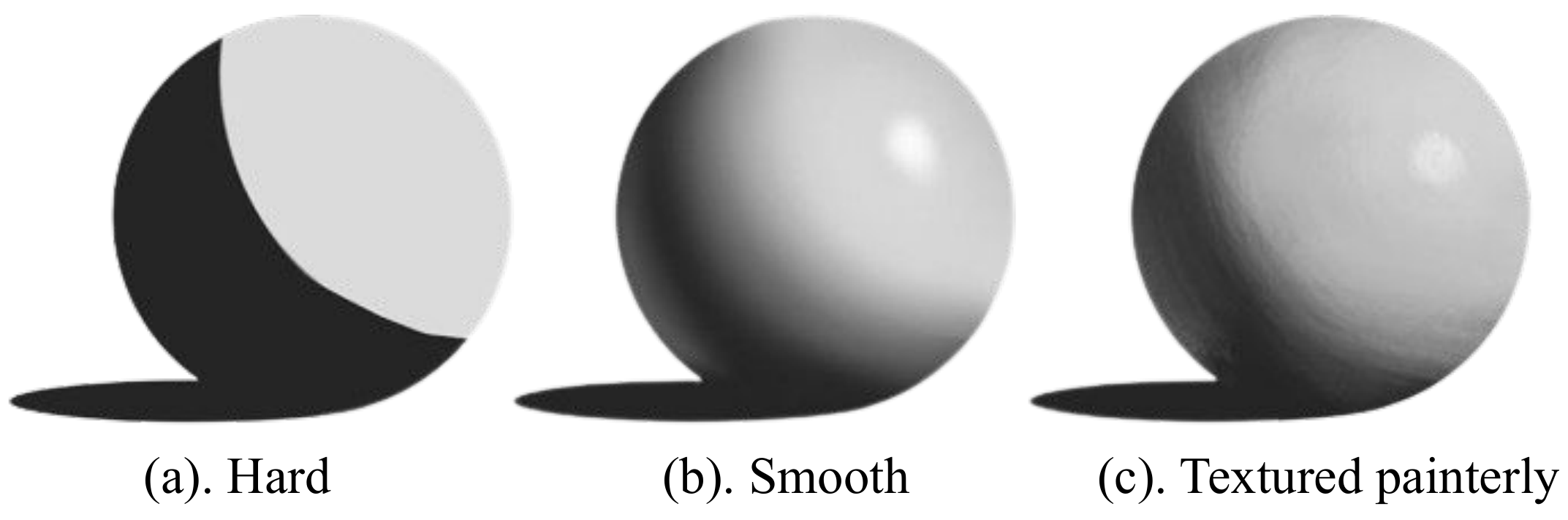} 
\caption{From left to right: (a) hard shading effects, (b) smooth shading effects, and (c) textured painterly shading effects.}    \label{fig:shading}
\end{figure}

Previous works on painting process generation have taken primarily two directions: painting video generation \cite{song2024processpainter, zhang2025generating, chen2024inverse} and painting stroke synthesis \cite{nakano2019neural, zou2021stylized, huang2019learning, liu2021paint, vinker2022clipasso, schaldenbrand2021styleclipdraw}. For painting video generation, the task is formulated as a video synthesis problem, where each frame represents a stage of the painting process. \cite{song2024processpainter} and \cite{chen2024inverse} explore diffusion models to predict time-lapse videos of artistic creation, while \cite{zhang2025generating} utilizes diffusion transformers to generate both past and future drawing processes. Although these works produce visually compelling videos thanks to powerful generative models, they do not provide explicit stroke information. Moreover, the transitions between frames are typically patch-based, making the generated painting workflows difficult to reproduce in real-world painting or existing digital painting software. In contrast, stroke-based painting generation methods explicitly represent brush strokes with parametric primitives, such as Bézier curves, to model individual stroke geometries and use differentiable renderers to synthesize vectorized images. Recent work has explored stroke parameter searching \cite{zou2021stylized}, reinforcement learning \cite{huang2019learning}, and transformer-based models \cite{liu2021paint} to generate brush stroke sequences. Nevertheless, neural painting approaches \cite{zou2021stylized, liu2021paint} often deviate from a realistic painting appearance. These methods render paint strokes defined by Bézier curves with single or dual colors on the canvas, lacking proper color blending between strokes. This limitation prevents them from producing complex textures, smooth tonal transitions, and natural shading effects characteristic of real paintings, resulting in hard stroke boundaries (Fig.~\ref{fig:shading} (a)) and an overall unrealistic visual appearance. Furthermore, when generating distinct painting styles such as oil, watercolor, or ink effects, these methods typically require training separate style-specific neural painters. This limitation restricts the generated outputs to a narrow set of predefined styles and hinders generalization beyond the training distribution.

Inspired by the progressive human painting process, which involves applying flat color fills followed by blending or smudging to create natural shading effects~\cite{jiang2024region}, we propose a differentiable stroke reconstruction framework that unifies stroke-based paint and smudge rendering with stylized texture generation, reproducing the paint-smudge loop to generate paintings with realistic shading and expressive textured strokes. Our pipeline is organized into three stages: (1) \textit{Paint Stroke Reconstruction}, a parallel differentiable color-filling module; (2) \textit{Stroke Texture Stylization}, a stylized texture synthesis module; and (3) \textit{Smudge Stroke Reconstruction}, a differentiable smudging operation for color blending. Unlike traditional painting software, where strokes are rasterized sequentially to match dynamically created user input, we introduce a parallel differentiable painting algorithm that renders single- or dual-color strokes directly from open Bézier curves in one step. With a differentiable paint renderer, we optimize Bézier-curve geometry parameters and per-stroke colors under geometric and semantic guidance. After extracting dual-color strokes and their geometric structures, we employ a StyleGAN-based generator to optimize latent textures conditioned on stroke geometry, producing diverse stylized appearances (e.g., watercolor, oil, and ink). The textured strokes are then composited onto the canvas and further refined through a differentiable smudge renderer, which generates smudged strokes and smooth color transitions to achieve natural shading. Emulating the iterative human painting process, we adopt a coarse-to-fine optimization strategy that progressively refines paint and smudge parameters by looping through painting, styling, and smudging, while spatially subdividing the canvas from coarse to fine patches. We evaluate our method on analog oil, watercolor, and ink paintings, as well as digital paintings, demonstrating visually compelling results with accurate reconstruction of stylized painting and smudge strokes.

In summary, our main contributions are as follows:
\begin{itemize}
    \item We propose a differentiable stroke reconstruction framework that unifies paint rendering, stylized texture generation, and smudge rendering, faithfully reproducing the paint–smudge loop with realistic and expressive strokes.
    \item We introduce a parallel differentiable stamp-based paint renderer and a differentiable smudge renderer that captures both geometry and texture, enabling joint optimization of stroke geometry, color, and texture under novel semantic and gradient guidance.
    \item We present a unified representation for stylized painting across diverse painting styles, and validate our approach through extensive experiments on various categories (oil, watercolor, ink, and digital paintings), demonstrating its effectiveness and expressiveness.
\end{itemize}

\section{Related Work}
\label{sec:related}

\begin{figure}[t]
\centering  
\includegraphics[width=0.45\textwidth]{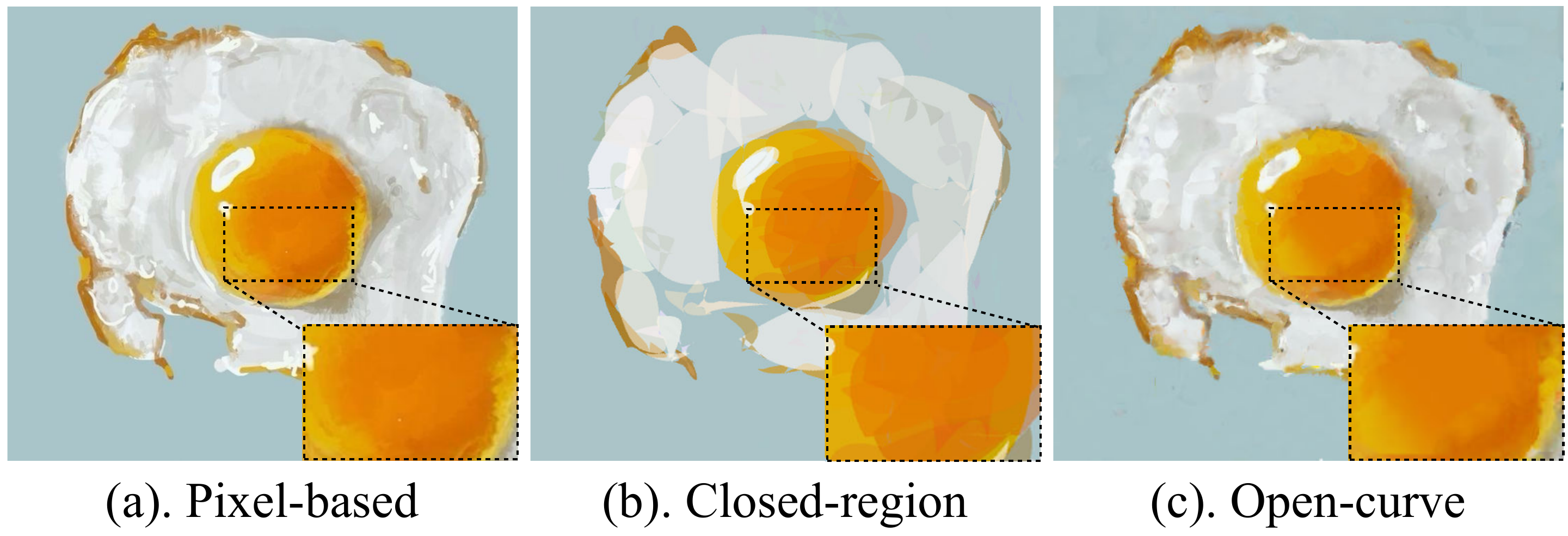} 
\caption{
From left to right: (a) input pixel-based image, (b) vectorized painting with closed-region representation, and (c) vectorized painting with open-curve representation. While the closed-region vectorization effectively captures fine details, its zigzag boundaries composed of multiple Bézier curves are less suitable for stroke-based painting and can only produce hard shading effects. In contrast, the open-curve representation models each stroke as a single cubic Bézier curve, blending colors through smudging. This approach more closely resembles human-created painting strokes and naturally produces smoother, more textured shading effects.}    \label{fig:closeopen}
\end{figure}

\subsection{Vector Graphics}
Vector graphics (VG) represents images parametrically as compositions of geometric primitives. Methods for generating SVGs fall into two broad families: data-driven and optimization-based. Data-driven approaches emit SVGs directly using sequence-to-sequence models \cite{reddy2021im2vec, lopes2019learned, wang2021deepvecfont}, diffusion models \cite{arar2025swiftsketch}, and transformers \cite{cao2023svgformer}, but learning-based methods are often limited to generating vector images within the distribution of training data, struggling with out-of-domain data \cite{liu2025b}. 

Optimization-based methods fit primitives by backpropagating through differentiable rasterizers to refine primitive parameters. DiffVG \cite{li2020differentiable} first introduced a differentiable vector rasterizer for vector graphics learning. DiffVG~\cite{li2020differentiable} first introduced a differentiable rasterizer for vector graphics learning and generation. Following this, \cite{ma2022towards, du2023image, hirschorn2024optimize, wang2025layered} have further explored layer-aware optimization with intersection-aware guidance~\cite{ma2022towards}, optimization in reduced subspaces with layer decomposition~\cite{du2023image}, top-down strategies with redundant stroke pruning~\cite{hirschorn2024optimize}, and semantic simplification methods~\cite{wang2025layered} to obtain semantically consistent SVGs. However, most optimization-based approaches are time-consuming~\cite{li2020differentiable,du2023image,ma2022towards}. To improve efficiency, Bézier Splatting \cite{liu2025b} leverages Bézier curves in combination with Gaussian splatting to achieve efficient vectorized image rendering~\cite{liu2025b}. Beyond image-to-VG tasks, differentiable vector rasterizers have been coupled with text guidance to synthesize vector drawings and sketches from prompts \cite{frans2022clipdraw, xing2023diffsketcher, schaldenbrand2021styleclipdraw, vinker2022clipasso, xing2024svgdreamer}. However, existing vector image generation explores closed, flat-filled color regions parameterized by Bézier curves \cite{li2020differentiable, ma2022towards, du2023image, hirschorn2024optimize, xing2023diffsketcher, vinker2022clipasso, xing2024svgdreamer}, lacking in creating smoothing shading effects. To achieve smooth-shaded vector images, diffusion curves \cite{orzan2008diffusion} and gradient meshes \cite{sun2007image} model gradient-based shading effects with topological control over the image in a single step. Unlike these global curve networks or control-mesh-based approaches, our goal is to explore open, stroke-based SVG primitives that can naturally reconstruct painting and shading effects accurately.

\subsection{Painting Video Generation}
With the advance of diffusion models, generation of high-quality images and videos has become possible \cite{rombach2022high, ho2022video, blattmann2023stable}. Prior work explores powerful generative models to synthesize the painting process as frame-by-frame sequences, effectively creating time-lapse videos that show how a painting emerges over time~\cite{chen2024inverse, zhang2025generating}. Inverse Painting~\cite{chen2024inverse} formulates the process as autoregressive image generation, leveraging image diffusion models with text and semantic guidance to produce each frame step by step. ProcessPainter~\cite{song2024processpainter} integrates image diffusion models with temporal attention to generate painting videos all at once. Going beyond forward processes, PaintsAlter~\cite{zhang2025generating} employs video diffusion models to achieve bidirectional painting process generation. Although these data-driven methods~\cite{chen2024inverse, zhang2025generating} generate high-quality painting videos, they often lack temporal consistency. In addition, the differences across frames typically appear as patchwise changes, lacking explicit stroke-level variations observed in human painting.

\subsection{Stroke-based Painting Generation}
Recently, \citet{nakano2019neural} proposed a neural painter with adversarial training to generate brushstroke paintings. Similarly, Learn to Paint \cite{huang2019learning} and Paint Transformer \cite{liu2021paint} explore reinforcement learning (RL) and a transformer-based framework, respectively, to learn the colors and positions of strokes. However, neural painting methods \cite{nakano2019neural, zou2021stylized, huang2019learning, liu2021paint} tend to use only one or two colors per stroke, lacking the ability to generate textured strokes. Thus, these methods often resort to an excessive quantity of tiny strokes to represent color in paintings. To produce stylized strokes, Stylized Neural Painting \cite{zou2021stylized} defined different stroke parameters and shapes for various painting styles, such as watercolor and oil painting. For each style, it was trained on different synthetic datasets and maintained separate weights. To generate a unified brushstroke style, Neural Brushstroke \cite{shugrina2022neural} leverages StyleGAN \cite{karras2019style} to extract latent style codes from input strokes, which can then be used in painting to create strokes with similar styles. Inspired by this, our framework aims to generate unified strokes that reconstruct both stroke geometry and stroke textures from a painting. Most prior stroke generation approaches \cite{nakano2019neural, zou2021stylized, huang2019learning, liu2021paint} focus only on color-filling strokes, lacking color blending, whereas human painting typically follows a paint–smudge loop in which colors are painted and then blended through blurring or smudging to create complex shading effects \cite{jiang2024region, shugrina2017playful}. Furthermore, neural renderers draw strokes all at once, lacking the stamp-based brushstroke rendering \cite{ciao2024ciallo} used in painting software. Our goal is to generate unified textured painting and smudging stamp-based brushstrokes that reproduce the human-like painting process and reconstruct natural smooth shading effects.



\begin{figure*}[t]
    \centering
        \includegraphics[width=\textwidth]{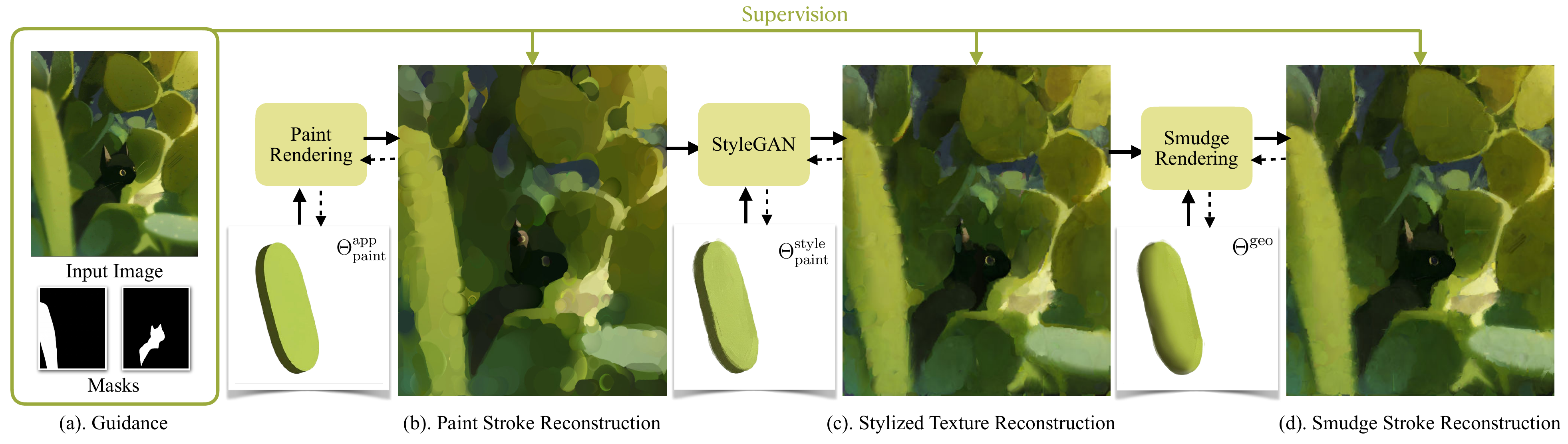}
        \vspace{-0.3in}
    \caption{\small{\textbf{Pipeline Overview.}
(a) Given a painting image as input, we extract segmentation masks and use both the RGB image and the masks as supervision for optimizing brush-stroke parameters.
(b) We then perform parallel differentiable paint rendering to rasterize strokes, with the geometric parameters optimized under appearance- and semantic-guided constraints.  
(c) Next, we employ a StyleGAN-based texture generator to synthesize stylized textured strokes conditioned on the geometric structures, optimizing the texture parameters of the paint strokes.  
(d) Finally, we apply a differentiable smudge renderer to simulate color blending and produce smooth shading effects, optimizing the smudge parameters.
}}\label{fig:pipeline}
\end{figure*} 

\section{Method}
\label{sec:method}


In this section, we discuss our reconstruction pipeline in depth.
Given an input painting, we progressively refine the canvas from coarse to fine by partitioning it into $1 \times 1$, $2 \times 2$, $\ldots$, $n \times n$ patches. 
Within each sub-canvas, our pipeline proceeds through three phases (see Fig.~\ref{fig:pipeline} for illustration):
(1) \textit{Paint stroke geometry reconstruction} recovers the geometric shapes of painting strokes under color guidance. 
(2) \textit{Paint stroke texture stylization} optimizes a style latent code conditioned on the reconstructed geometry to generate diverse stroke textures. 
(3) \textit{Smudge stroke reconstruction} optimizes the shapes of smudging strokes to reconstruct smooth shading effects. 
These three phases are applied iteratively from coarse to fine canvas levels, progressively refining the reconstructed painting and emulating the human paint–smudge loop.

\subsection{Stroke Representation}  
To support the three-phase pipeline described above, we adopt a unified representation for both paint and smudge strokes. Each stroke is parameterized as $\{\mathbf{x}_{B}, \mathbf{r}, \mathbf{c}, \alpha, \mathbf{w}\}$, where $\mathbf{x}_{B}=\{\mathbf{x}_s,\mathbf{x}_e,\mathbf{x}_c\}$ are Bézier curve endpoints and control points, $\mathbf{r}=\{r_s,r_e\}$ are endpoint radius, $\mathbf{c}=\{\mathbf{c}_s,\mathbf{c}_e\}$ are endpoint colors, $\alpha$ is transparency, and $\mathbf{w}$ is a style latent code controlling texture and appearance. 
We decompose the parameters into geometry $\Theta^{\text{geo}}=\{\mathbf{x}_{B}, \mathbf{r}\}$, color $\Theta^{\text{color}}=\{\mathbf{c}, \alpha\}$, and style $\Theta^{\text{style}}=\{\mathbf{w}\}$ components.
For paint strokes, we separately optimize their appearance (including geometry and color) $\Theta_{\text{paint}}^{\text{app}}=\{\mathbf{x}_{B}, \mathbf{r}, \mathbf{c}, \alpha\}$ and style components $\Theta_{\text{paint}}^{\text{style}}=\{\mathbf{w}\}$. For smudge strokes, we optimize only the geometry parameters $\Theta_{\text{smudge}}=\{\mathbf{x}_{B}, \mathbf{r}\}$.


\begin{algorithm}[t]
\caption{Stroke-Based Paint Rendering}
\begin{algorithmic}[1]
\Require Stroke parameters $\Theta_{\text{paint}} = \{\mathbf{x}_B = (\mathbf{x}_s, \mathbf{x}_c, \mathbf{x}_e), \mathbf{r} = (r_s, r_e), \mathbf{c} = (\mathbf{c}_s, \mathbf{c}_e), \alpha\}$, number of stamps $N$
\Ensure Rendered pixel-based stroke $\bm{\tau}_p$

\State $\mathcal{S} \gets \emptyset$
\Comment{\textbf{Initialize stamp}}
\For{$k = 0$ to $N$}
    \State $t_k \gets \frac{k}{N}$
    \State $\mathbf{x}_k \gets (1 - t_k)^2 \mathbf{x}_s + 2 t_k (1 - t_k) \mathbf{x}_c + t_k^2 \mathbf{x}_e$
    \State $r_k \gets (1 - t_k) r_s + t_k r_e$
    \State $\mathbf{c}_k \gets (1 - t_k) \mathbf{c}_s + t_k \mathbf{c}_e$
    \State $S_k(\mathbf{x}) = \alpha\,\mathbf{c}_k$ if $\|\mathbf{x} - \mathbf{x}_k\| \leq r_k$, else $\mathbf{0}$
    \State $\mathcal{S} \gets \mathcal{S} \cup \{S_k\}$
\EndFor


\ForAll{pixels $\mathbf{x}$} \Comment{\textbf{SDF-based rendering}}
    \State $k^* \gets \arg\min_k \|\mathbf{x} - \mathbf{x}_k\|$
    \State $p(\mathbf{x}) \gets \alpha \cdot \mathbf{c}_{k^*}$
\EndFor

\State \Return $\bm{\tau}_p = \{p(\mathbf{x})\}$
\end{algorithmic}
\end{algorithm}

\subsection{Differentiable Stroke Rendering}

\subsubsection{Stroke-based paint rendering}  
\begin{figure}[t]
\centering  \includegraphics[width=0.45\textwidth]{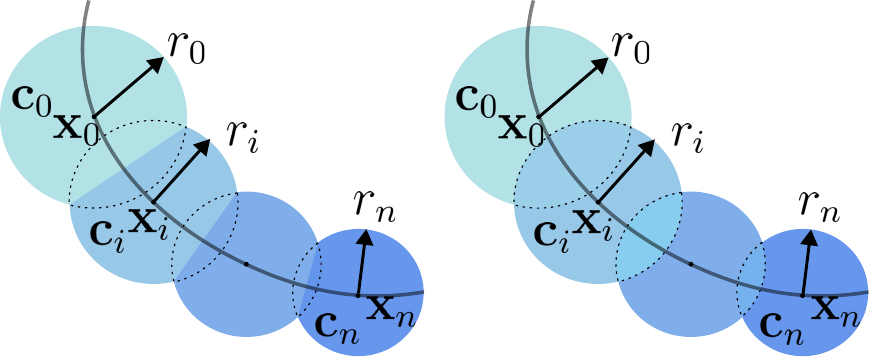} 
\caption{\textbf{Differentiable Paint Renderer.} Local stamps are uniformly sampled along the Bézier curve and parameterized by radius, color, and transparency. Unlike the vanilla stroke-based renderer (right), where interpolated areas are sequentially blended using the colors of neighboring stamps, we propose an distance-based model in which colors are determined by the closest stamp centers (left), allowing efficient parallelization.}    \label{fig:diffpaint}
\end{figure}

Given the paint stroke parameters $\Theta_{\text{paint}}^{\text{app}} = \{\mathbf{x}_{B}, \mathbf{c}, \mathbf{r}, \alpha\}$, our differentiable paint renderer $\mathbf{\phi}_p$ produces the corresponding pixel-based stroke $\bm{\tau}_p = \phi_p(\Theta_{\text{paint}}^{\text{app}})$ (Fig. \ref{fig:diffpaint}). Following standard stamp-based painting algorithms~\cite{ciao2024ciallo}, a stroke is synthesized by placing a sequence of stamps along the Bézier path using parameter-uniform sampling. The stamp centers are computed by uniformly sampling the Bézier curve at $\mathbf{x}_k = (1-t_k)^2\mathbf{x}_s + 2t_k(1-t_k)\mathbf{x}_c + t_k^2\mathbf{x}_e$, where $t_k = \frac{k}{N} (k=0,\ldots,N)$,
with radius interpolated as $r_k=(1-t_k)r_s+t_k r_e$ and color interpolated as 
$\mathbf{c}_k=(1-t_k)\mathbf{c}_s+t_k\mathbf{c}_e$.
Finally, we define a local stamp centered at $\mathbf{x}_k$, parameterized by its radius $r_k$, 
color $\mathbf{c}_k$, and transparency $\alpha$:
\begin{equation}
S_k(\mathbf{x}) = 
S(\mathbf{x}-\mathbf{x}_k; r_k, \mathbf{c}_k, \alpha) =
\begin{cases}
\alpha\,\mathbf{c}_k, & \|\mathbf{x}-\mathbf{x}_k\|\le r_k, \\[2pt]
\mathbf{0}, & \text{otherwise}.
\end{cases}
\end{equation}
For each image pixel $\mathbf{x}=(x,y)$, its color is obtained by 
accumulating the contributions of all stamps whose support covers $\mathbf{x}$, 
i.e., those satisfying $\|\mathbf{x}-\mathbf{x}_k\|\leq r_k$, combined through 
sequential alpha compositing:
\begin{equation}
p(\mathbf{x}) = \sum_{k=0}^N 
\Big[ S_k(\mathbf{x})
\prod_{j<k}\bigl(1-\alpha\,\mathbf{1}[\|\mathbf{x}-\mathbf{x}_j\|\le r_j]\bigr)\Big].
\end{equation}
To avoid sequential accumulation, we also introduce a signed distance function (SDF) representation where we evaluate:
\begin{equation}
d(\mathbf{x})=\min_{k}\; \bigl(\|\mathbf{x}-\mathbf{x}_k\|-r_k\bigr),
\end{equation}
which implicitly defines the stroke region by the zero-level set $\{\mathbf{x}:d(\mathbf{x})\leq 0\}$. Pixel colors are then computed in parallel by assigning each pixel to its nearest stamp center:
\begin{equation}
p(\mathbf{x})=\mathbf{c}_{k^*}\,\alpha, \qquad 
k^*=\arg\min_{k}\|\mathbf{x}-\mathbf{x}_k\|,
\end{equation}
where $p(\mathbf{x})$ computes the pixel value of the rendered paint stroke $\bm{\tau}_p$ at location $\mathbf{x}$. This approach removes the need for sequential blending and allows efficient stroke construction on GPUs.

\begin{algorithm}[t]
\caption{Stroke-Based Smudge Rendering}
\begin{algorithmic}[1]
\Require Smudge parameters $\Theta_{\text{smudge}} = \{\mathbf{x}_B = (\mathbf{x}_s, \mathbf{x}_c, \mathbf{x}_e), \mathbf{r} = (r_s, r_e)\}$, canvas $I \in [0,1]^{3 \times H \times W}$, number of stamps $N$
\Ensure Smudged stroke $\bm{\tau}_s$

\State Sample trajectory $\{\mathbf{x}_0, \dots, \mathbf{x}_N\}$ along Bézier curve
\For{$k = 0$ to $N$}
\Comment{\textbf{Initialize stamp}}
    \State $t_k \gets \frac{k}{N}$
    \State $\mathbf{x}_k \gets (1 - t_k)^2 \mathbf{x}_s + 2 t_k (1 - t_k) \mathbf{x}_c + t_k^2 \mathbf{x}_e$
    \State $r_k \gets (1 - t_k) r_s + t_k r_e$
    \State $S_k(\mathbf{x}) \gets \{ I(\mathbf{x}) \mid \|\mathbf{x} - \mathbf{x}_k\| \leq r_k \}$
\EndFor

\State $C_k^0(\mathbf{x}), B_k(\mathbf{x}) \gets S_k(\mathbf{x})$ for $k = 0,\dots,N $


\For{$k = 0$ to $N$} \Comment{\textbf{Brush update}}
    \State $t_k \gets \ell_k / L$
    \Comment{ratio of arc length to total length}
    \State
    \(
    \mathcal{K}_{k,i} = \frac{t_i^{a-1}(1 - t_i)^{b-1}}{\sum_{m=0}^{k} t_m^{a-1}(1 - t_m)^{b-1}} \quad \text{for } i = 0,\dots,k
    \)
    \State $B_k(\mathbf{x}) \gets \sum_{i=0}^k \mathcal{K}_{k,i} \cdot B_i(\mathbf{x})$
\EndFor

\For{$k = 1$ to $N$} \Comment{\textbf{Canvas update}}
    \State $C_k^k(\mathbf{x}) \gets \alpha_c B_{k-1}(\mathbf{x}) + (1 - \alpha_c) C_k^{k-1}(\mathbf{x})$
    \State $B_k^k(\mathbf{x}) \gets \alpha_c B_k^{0}(\mathbf{x}) + (1 - \alpha_s) C_k^{k}(\mathbf{x})$
\EndFor

\State \Return $\bm{\tau}_s = \{C_N^N(\mathbf{x})\}$

\end{algorithmic}
\end{algorithm}

\subsubsection{Stroke-based smudge rendering}

\begin{figure}[t]
\centering  \includegraphics[width=0.45\textwidth]{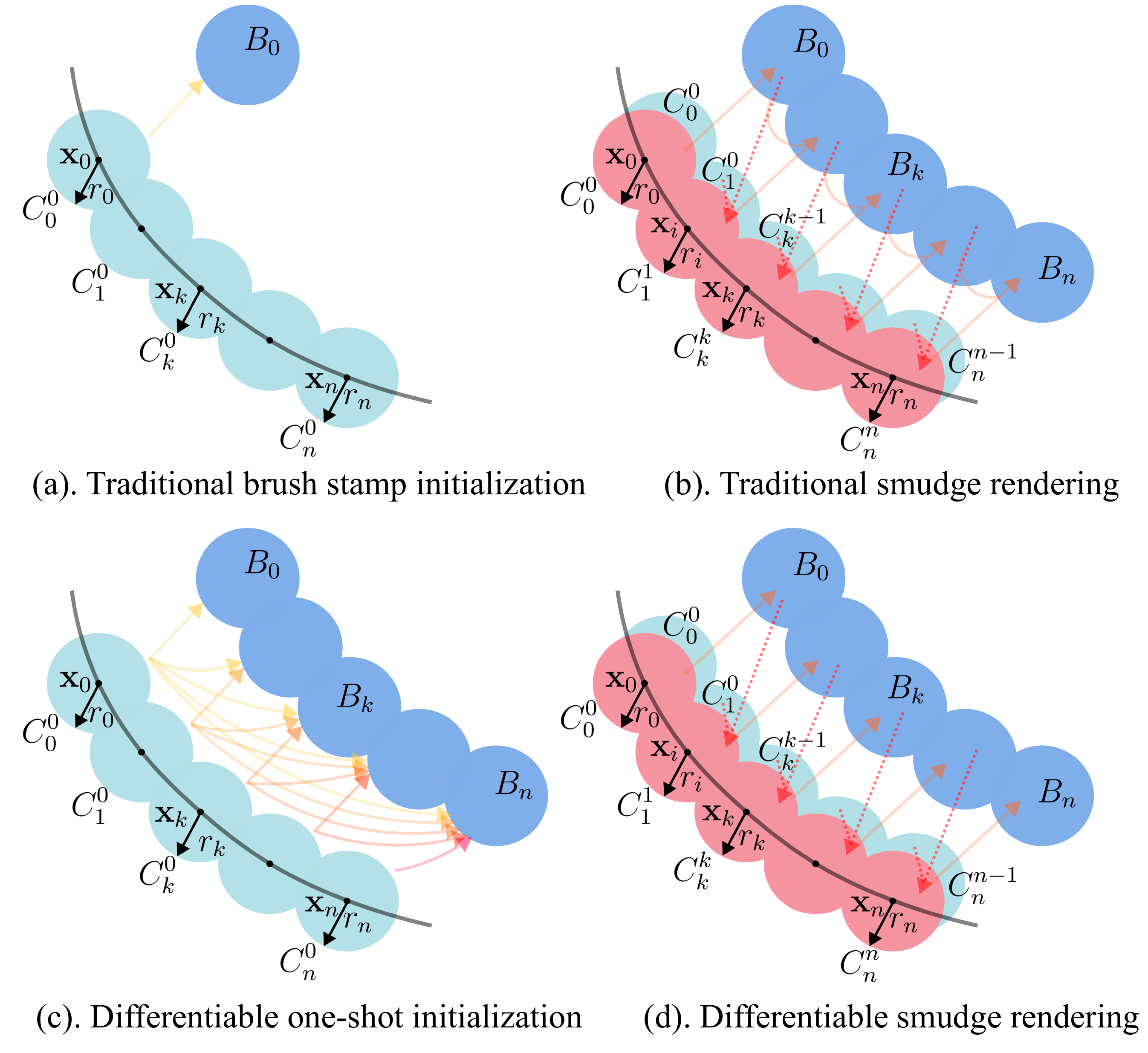} 
\caption{\textbf{Differentiable smudge renderer.} Blue, cyan, and red circles represent the brush states and the initial and updated canvas states. In traditional smudge rendering, each brush stamp is initialized sequentially (a) and alternately updates the canvas and brush states in a recursive manner (b). In contrast, our proposed differentiable smudge renderer introduces a novel one-shot initialization that updates all brush states simultaneously (c) and updates brush stamps directly from the canvas, resulting in a more efficient computation graph (d).}    \label{fig:diffsmudge}
\end{figure}

Given the smudge stroke parameters  $\Theta_{\text{smudge}}=\{\mathbf{x}_{B}, \mathbf{r}\}$ 
and an input canvas $I\in[0,1]^{3\times H\times W}$, our differentiable smudge renderer $\phi_s$ produces the corresponding 
pixel-based smudge stroke: $\bm{\tau}_s = \phi_s(\Theta_{\text{smudge}}, I).$ The smudge trajectory is obtained by uniformly sampling the Bézier curve as 
$\{\mathbf{x}_0,\ldots,\mathbf{x}_N\}$, where each location $\mathbf{x}_k$ 
is associated with an interpolated radius $r_k$. 
For each $\mathbf{x}_k$, we obtain a local stamp $S_k(\mathbf{x})$ by extracting a patch of size $r_k$ centered at $\mathbf{x}_k$ from the input canvas $I$. 


Traditional smudging evolves the canvas state $C$ and the brush state $B$ in a recursive Markovian manner. Let $B_k$ denote the internal brush state of stamp $k$, and $C_k^n$ denote the updated canvas state of stamp $k$ at temporal step $n$. The canvas and brush states evolve following the \textit{alternating update} scheme (Fig.~\ref{fig:diffsmudge} (b)):
\begin{align}
C_k^k(\mathbf{x}) &= \alpha_c B_{k-1}(\mathbf{x}) + (1-\alpha_c) C_k^{k-1}(\mathbf{x}), \\
B_k(\mathbf{x}) &= \alpha_s B_{k-1}(\mathbf{x}) + (1-\alpha_s) C_k^k(\mathbf{x}),
\end{align}
with initialization \(B_0(\mathbf{x})=C_0^0(\mathbf{x})\) (Fig.~\ref{fig:diffsmudge} (a)), where \(\alpha_c\in[0,1]\) is the canvas blending coefficient controlling how much of the previous brush state remains on the canvas, and \(\alpha_s\in[0,1]\) is the self-retention coefficient controlling how much pigment the brush retains from its own past state during the smudging process. However, since each step extracts and splits stamps directly from the canvas, this formulation is difficult to make differentiable. Moreover, each current stamp depends only on the previous stamp and the current canvas state, resulting in a shallow computation graph with only one-step temporal dependencies.

To overcome these limitations, we note that the recurrence can be simplified as \(B_k = \mathcal{A} B_{k-1} + \mathcal{B} S_k\), where \(\mathcal{A}=\alpha_s+(1-\alpha_s)\alpha_c\), \(\mathcal{B}=(1-\alpha_s)(1-\alpha_c)\),
and unrolled to
\[
B_k = \mathcal{A}^k C_0^0 + \sum_{i=1}^k \mathcal{A}^{k-i}\mathcal{B}\,C_i^{i-1},
\]
which shows that past samples \(\{C_i\}\) contribute with exponentially decaying weights. Inspired by this, we propose a one-shot initialization with a length-aware distribution to generate $B_k$ given initial canvas stamps (Fig. \ref{fig:diffsmudge} (c)). Reparameterizing the trajectory by cumulative
arc length $\ell_i$ with total length $L$ and normalized positions
$t_i=\ell_i/L$, we obtain
\[
B_k = \sum_{i=0}^k \mathcal{K}_{k,i}\,C_i^0,\quad
\mathcal{K}_{k,i} = \frac{t_i^{\,a-1}(1-t_i)^{\,b-1}}
{\sum_{m=0}^{k} t_m^{\,a-1}(1-t_m)^{\,b-1}},
\]
where $a,b>0$. A special case $a=b=1$ recovers the uniform kernel $\mathcal{K}_{k,i}=1/(k+1)$, corresponding to the arithmetic mean of all past samples.
In addition, the kernel weights $\mathcal{K}_{k,i}$ capture arc-length geometry and remain invariant to resampling density. This property enables the precomputation of the matrix $\mathcal{K}$ and the one-shot initialization of all brush states $\{B_k\}$. After initializing brush stamps, we use it to update canvas states and adjust brush stamps from only canvas states recursively (Fig. \ref{fig:diffsmudge} (d)):
\begin{align}
C_k^k(\mathbf{x}) &= \alpha_c B_{k-1}(\mathbf{x}) + (1-\alpha_c) C_k^{k-1}(\mathbf{x}),\\
B_k^k(\mathbf{x}) &= \alpha_s B_k^{0}(\mathbf{x}) + (1-\alpha_s) C_k^k(\mathbf{x}).
\end{align}
This ordered blending ensures pigment deposition aligns with the temporal smudging process.



\subsection{Stroke Reconstruction}

\subsubsection{Phase I: Paint stroke appearance reconstruction}\label{sec:paint_geo_recon}

At this stage, we optimize only the appearance parameters of each paint stroke, $\Theta_{\text{paint}}^{\text{app}}$, while keeping the style latent $\mathbf{w}$ fixed for texture generation in a later stage.
To guide the optimization, we employ a combination of complementary objectives.
\paragraph{Appearance Alignment}
We employ a straightforward pixel loss, $\mathcal{L}_{\text{pixel}}=\|I_r-I_t\|_{1}$, to minimize the discrepancy between the rendered image $I_r$ and the ground truth $I_t$ in raw color space, encouraging faithful reconstruction in RGB space.
In addition, we incorporate a perceptual loss \cite{shugrina2022neural}, defined as $\mathcal{L}_{\text{perc}}=\sum_{\ell}\|F_{\ell}(I_r)-F_{\ell}(I_t)\|_{1}$, where $F_\ell(\cdot)$ denotes the feature map extracted from the $\ell$-th layer of a pretrained VGG network \cite{johnson2016perceptual}. This loss captures both low-level cues (e.g., edges and colors) and high-level semantic patterns, providing perceptual guidance beyond raw pixel differences.

\paragraph{Structural Guidance}
To provide further structural guidance for the strokes, we utilize the gradient information of the image and propose the following loss:
a gradient-based alignment loss $\mathcal{L}_{\text{grad}}=\tfrac{1}{L}\sum_{\ell}(\alpha\,\mathcal{L}_{\text{mag}}^{(\ell)}+\beta\,\mathcal{L}_{\text{dir}}^{(\ell)})$, where $L$ is the total arc length, $\mathcal{L}_{\text{mag}}$ and $\mathcal{L}_{\text{dir}}$ measure the difference in gradient magnitudes and orientations between $I_r$ and $I_t$, respectively. This loss encourages the strokes to align with local geometry such as edges and contours. 
Additionally, to ensure each stroke aligns with the intended object region and to prevent strokes from crossing multiple objects and causing undesired blending, we introduce a layer-based segmentation loss, $\mathcal{L}_{\text{seg}} = \sum_{i \ne i^*} |A_s \cap M_i|$, where $A_s$ is the predicted stroke mask and ${M_i}$ are the semantically segmented regions, extracted by Segment Anything (SAM) \cite{kirillov2023segment}. The index $i^* = \arg\max_i |A_s \cap M_i|$ corresponds to the intended object region.

\paragraph{Optimization Regularization}
To avoid vanishing gradients when strokes are far from their targets, we adopt an entropy-regularized optimal transport loss \cite{zou2021stylized}: 
$$\mathcal{L}_{\text{OT}} = \left\langle C, \arg\min_{P \in \mathcal{U}} \left\langle C, P \right\rangle - \frac{1}{\lambda} \left( -\sum_{i,j=1}^{n} P_{i,j} \log P_{i,j} \right) \right\rangle,$$
which matches mass distributions between $I_r$ and $I_t$ and stabilizes optimization.  
Finally, an area regulation loss:  
$$\mathcal{L}_{\text{area}}=\tfrac{1}{M}\sum_{s=1}^M \exp(-\tfrac{\mathrm{area}(A_s)}{\eta})$$  
penalizes strokes with vanishing areas, ensuring that each stroke maintains a minimal effective footprint.  

The overall objective is a weighted combination of all these terms:  
\begin{multline}
\mathcal{L}^{\text{app}}=
\lambda_{\text{pixel}}\mathcal{L}_{\text{pixel}}+
\lambda_{\text{perc}}\mathcal{L}_{\text{perc}}+\\
\lambda_{\text{grad}}\mathcal{L}_{\text{grad}}+
\lambda_{\text{seg}}\mathcal{L}_{\text{seg}}+
\lambda_{\text{OT}}\mathcal{L}_{\text{OT}}+
\lambda_{\text{area}}\mathcal{L}_{\text{area}},
\end{multline}
where $\lambda_{\ast}$ are scalar weights balancing the contributions of each loss.

\subsubsection{Phase II: Stylized Texture Reconstruction}  
After obtaining the optimized appearance parameters $\Theta_{\text{paint}}^{\text{app}}$, we keep them fixed and optimize the style latent code $\mathbf{w}$. Optimizing $\mathbf{w}$ allows us to synthesize diverse brushstroke textures (e.g., oil, watercolor, ink, and digital painting styles). We adopt a conditional StyleGAN generator~\cite{shugrina2022neural}, denoted as $G$, to produce a stylized textured paint stroke $
\bm{\tau}_{st} = G(\Theta_{\text{paint}}^{\text{app}}, \mathbf{w})$, where $\mathbf{w}$ controls texture and appearance variations. During optimization, only $\mathbf{w}$ is updated, ensuring that the geometry remains unchanged.

To guide stylization, we apply selective reconstruction losses from Section~\ref{sec:paint_geo_recon}, specifically, the appearance alignment losses in RGB and latent spaces, as well as the gradient loss for localized structural preservation. These losses enforce color fidelity, structural similarity, and consistent shading, while the rest of the geometric and regularization terms are excluded, as the stroke shape is fixed. The overall objective for the stylization stage is the following:
\[
\mathcal{L}^{\text{style}} = 
\lambda_{\text{pixel}}\mathcal{L}_{\text{pixel}} +
\lambda_{\text{perc}}\mathcal{L}_{\text{perc}} +
\lambda_{\text{grad}}\mathcal{L}_{\text{grad}}.
\]

\subsubsection{Phase III: Smudge stroke reconstruction}  
At this stage, we optimize the trajectories of smudge strokes. Each smudge stroke is parameterized by $\Theta_{\text{smudge}}=\{\mathbf{x}_{B},\mathbf{r}\}$. We use the same reconstruction loss $\mathcal{L}^{\text{app}}$ as described in Section~\ref{sec:paint_geo_recon}, since our optimization objectives remain the same. The primary difference lies in the differentiable renderer employed during each stage. Additionally, we assign greater weights to the gradient magnitude loss $\mathcal{L}_{\text{mag}}$ to provide stronger shading guidance, as well as the area regularization loss $\mathcal{L}_{\text{area}}$ to prevent degenerate stroke configurations.


\subsection{Joint Optimization}  
Inspired by the iterative human paint–smudge process, we adopt a coarse-to-fine optimization strategy. The canvas is hierarchically partitioned, starting from a coarse $1 \times 1$ grid and progressively subdividing into finer grids: $2 \times 2$, $\cdots$, up to $n \times n$ sub-canvases. At each level, we first reconstruct painting strokes with both appearance and stylization, followed by smudge stroke optimization to refine shading and blending. The output from each level initializes the next, enabling gradual refinement of details while preserving the global structure. At the final $n \times n$ level, however, we perform only paint-stroke reconstruction, omitting smudge strokes to maintain edge sharpness and avoid over-smoothing.

\section{Experiments}
\label{sec:exp}

\subsection{Implementation Details}  
We implement our algorithm in PyTorch~\cite{paszke2019pytorch}. For paint stroke appearance reconstruction and smudge stroke reconstruction, we use RMSprop with a learning rate of $0.003$. For stylized textured stroke reconstruction, we employ the Adam optimizer with a learning rate that linearly warms up from 0 to 0.01 over the first 5\% of iterations, remains constant for the next 70\%, and then follows a cosine decay schedule to 0 over the final 25\%. The weights of each loss term are set as follows:
$\lambda_{\text{pixel}}=1.0$, $\lambda_{\text{seg}}=0.1$, $\lambda_{\text{area}}=0.02$, $\lambda_{\text{grad}}=0.1$, $\lambda_{\text{OT}}=0.2$, and $\lambda_{\text{perc}}=0.1$. 
For stroke texture rendering, we use a pretrained StyleGAN generator~\cite{shugrina2022neural} trained on $128\times128$ patches. Following the initialization strategy of Stylized Neural Painting~\cite{zou2021stylized}, we construct an error map by comparing the target image with the current canvas prediction. Specifically, we compute the $\mathcal{L}^1$ distance per pixel, summed across the RGB channels, to guide stroke placement toward regions with higher reconstruction error. All experiments are conducted on a single NVIDIA H100 GPU.

\subsection{Stroke Rendering Results}

\begin{figure}[t]
    \centering
    \includegraphics[width=\columnwidth]{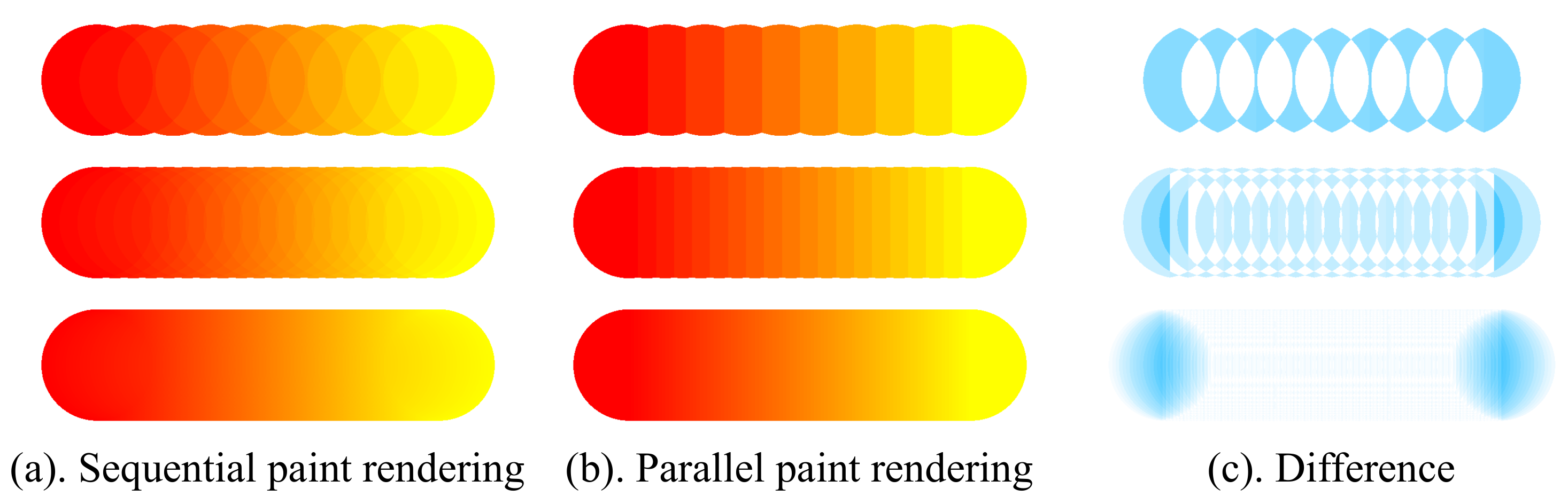}
    \vspace{-2mm}
    \caption{
         \textbf{Sequential vs. parallel differentiable paint rendering.} With 10, 20, and 100 strokes (from top to bottom), we compare (a) sequential rendering and (b) our proposed parallel rendering. The results show that the parallel rendering achieves accuracy comparable to the sequential approach. (c) The difference visualization highlights regions of discrepancy, where brighter blue indicates larger differences.
    }
    \label{fig:paintres}
    \vspace{-2mm}
\end{figure}

\paragraph{Paint Stroke Rendering.}
We evaluate our proposed parallel differentiable paint stroke renderer with sampling densities of 10, 20, and 100 stamps on a $1024 \times 1024$ canvas using a stroke radius of 100 pixels, and compare its performance against the traditional sequential stroke-based rendering method. As shown in Fig.~\ref{fig:paintres}, our method maintains accurate rendering results across different sampling densities. To quantitatively assess efficiency and accuracy, we benchmark the proposed parallel stroke renderer against the sequential Bézier stroke renderer under identical settings on an NVIDIA H100 GPU. For each test stroke, we record the forward-pass time averaged over five runs, with CUDA synchronization before and after each iteration to ensure that only GPU computation time is measured, excluding any I/O or visualization overhead. We report both the average rendering time and the $\mathcal{L}^1$ distance per pixel between the two outputs. Our parallel implementation achieves up to a 7$\times$ speedup on 100-stamp strokes while maintaining comparable rendering fidelity (see Table~\ref{tab:seq_par_l1}).

\begin{table}[htbp]
\centering
\footnotesize
\caption{Comparison between sequential and parallel paint rendering.}
\label{tab:seq_par_l1}
\begin{tabular}{lcccc}
\hline
\textbf{\# Stamp} & \textbf{Seq. Time(ms)} & \textbf{Par. Time(ms)} & \textbf{Acc. Ratio} $\uparrow$ & \textbf{$\mathcal{L}^1$ Dist.} $\downarrow$ \\
\hline
10   & 2.335 & 0.912 & 2.56$\times$ & 0.003171 \\
20   & 3.980 & 1.033 & 3.85$\times$ & 0.002896 \\
100  & 17.532 & 2.424 & 7.23$\times$ & 0.002177 \\
\hline
\end{tabular}
\end{table}

\begin{figure}[t]
    \centering
    \includegraphics[width=\columnwidth]{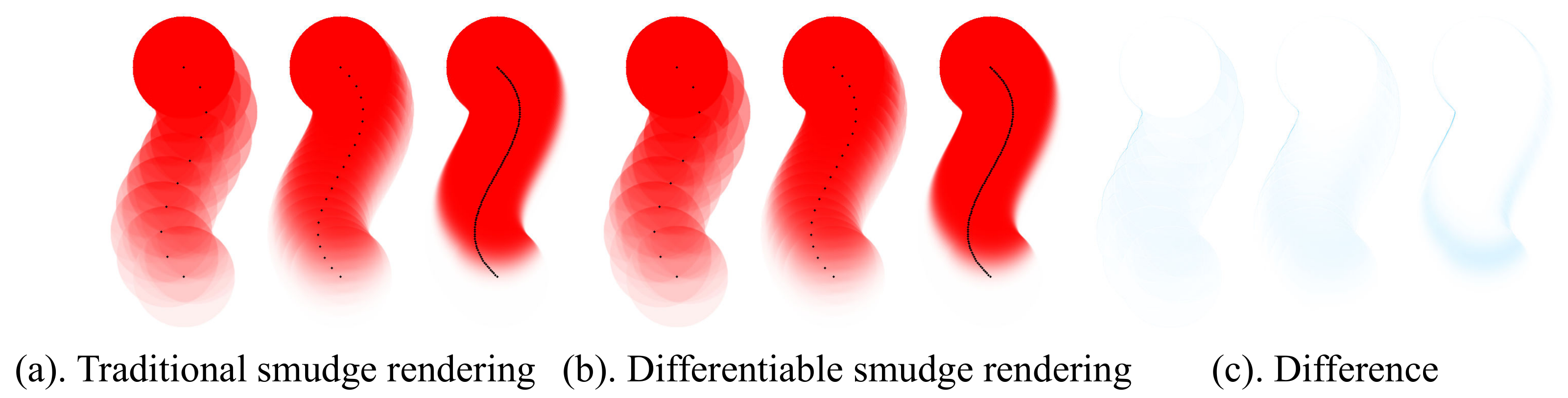}
    \vspace{-2mm}
    \caption{
         \textbf{Traditional vs. differentiable smudge rendering.} With 10, 20, and 100 strokes (from left to right), the black dots indicate smudge trajectories, and the input is a single red circle. Comparing (a) traditional smudge rendering with (b) our differentiable smudge rendering using one-shot initialization, we observe that the proposed parallel rendering achieves accuracy comparable to the traditional method. In (c), the difference visualization shows that brighter blue regions correspond to larger differences.
    }
    \label{fig:smudgeres}
    \vspace{-2mm}
\end{figure}

\paragraph{Smudge Stroke Rendering.}
We further evaluate the proposed differentiable smudge stroke renderer using sampling densities of 10, 20, and 100 stamps on a $1024 \times 1024$ canvas with a stroke radius of 100 pixels, and compare its performance against a vanilla stroke-based smudge rendering baseline. As the number of samples per stroke increases, the shading becomes progressively smoother. As shown in Fig.~\ref{fig:smudgeres}, our proposed differentiable smudge renderer with one-shot initialization achieves accurate and stable smudge rendering results.

\subsection{Full Reconstruction Results}
Fig.~\ref{fig:teaser} showcases reconstructed paintings generated with 1024 strokes across four representative styles, including watercolor, oil, digital, and ink, demonstrating that our method effectively reproduces vivid colors, smooth brush transitions, and faithful stylistic details. We refer readers to the supplementary document for extensive painting reconstruction results.

\subsection{Applications}
\paragraph{Layer-Aware Painting Generation.}
Layer decomposition has been widely used in image editing~\cite{tan2016decomposing, yang2025generative} and vectorization tasks~\cite{ma2022towards}. In our brushstroke reconstruction framework, we further incorporate layer decomposition into the painting generation pipeline, enabling a \emph{layer-aware generative painting process} that produces semantically structured results and facilitates subsequent layer-based editing.

\begin{figure}[t]
    \centering
    \includegraphics[width=\columnwidth]{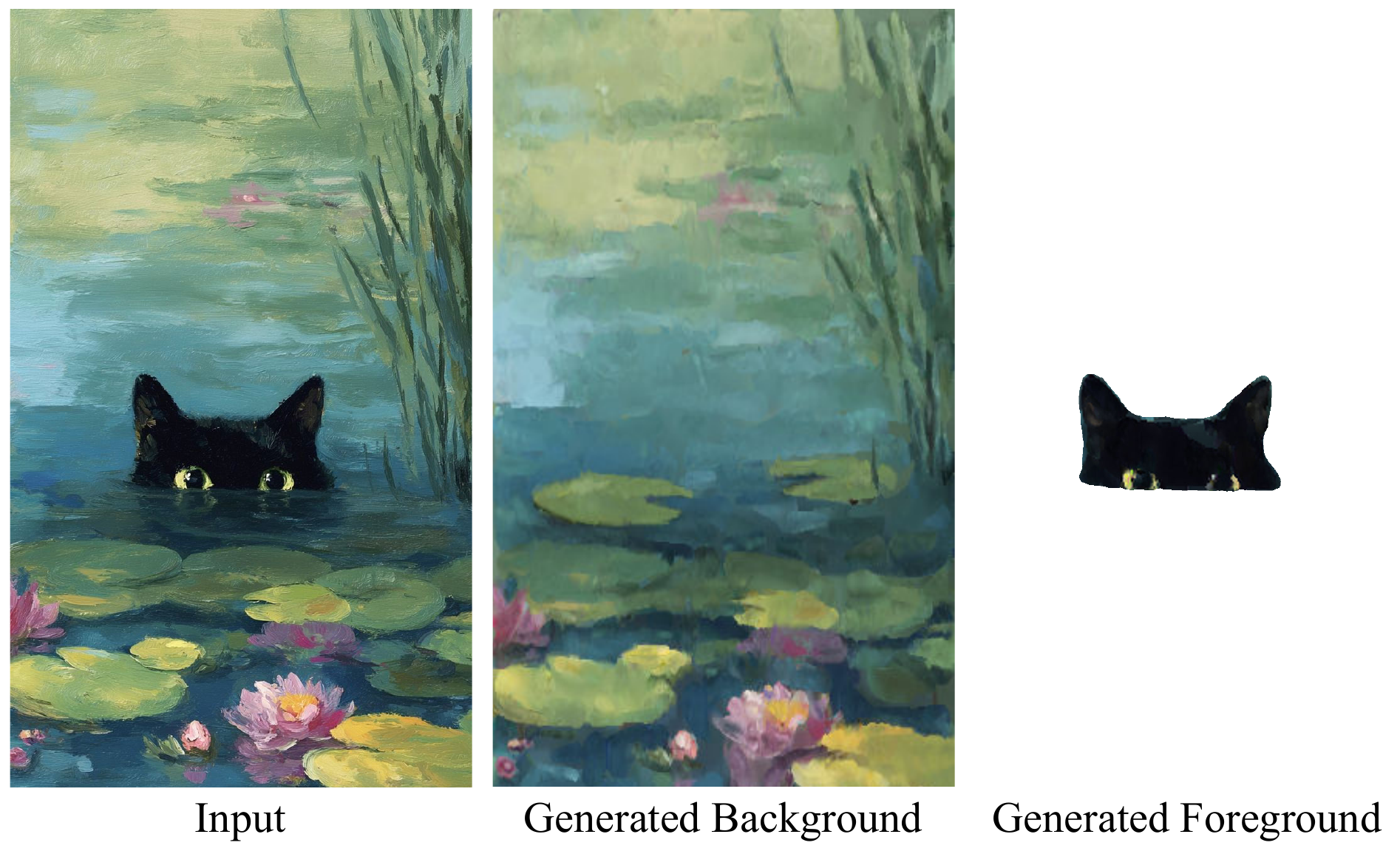}
    \vspace{-2mm}
    \caption{\textbf{Layered Painting Generation.} Our method enables high-quality layer-aware painting generation from a single image.}
    \label{fig:layer}
    \vspace{-2mm}
\end{figure}

To generate decomposed painting layers, we first employ semantic segmentation model ~\cite{ravi2024sam, liu2024grounding} to separate foreground objects from the background. Since transparent and semi-transparent regions often contain mixed visual content, we further use a vision–language model~\cite{team2023gemini} to inpaint and refine the background layer. Subsequently, stroke reconstruction is performed within each segmented region, resulting in a hierarchical painting generation process that preserves both semantic separation and artistic coherence. As shown in Fig.~\ref{fig:layer}, the proposed approach produces realistic and editable paintings with consistent textures across layers.

\begin{figure}[t]
    \centering
    \includegraphics[width=\columnwidth]{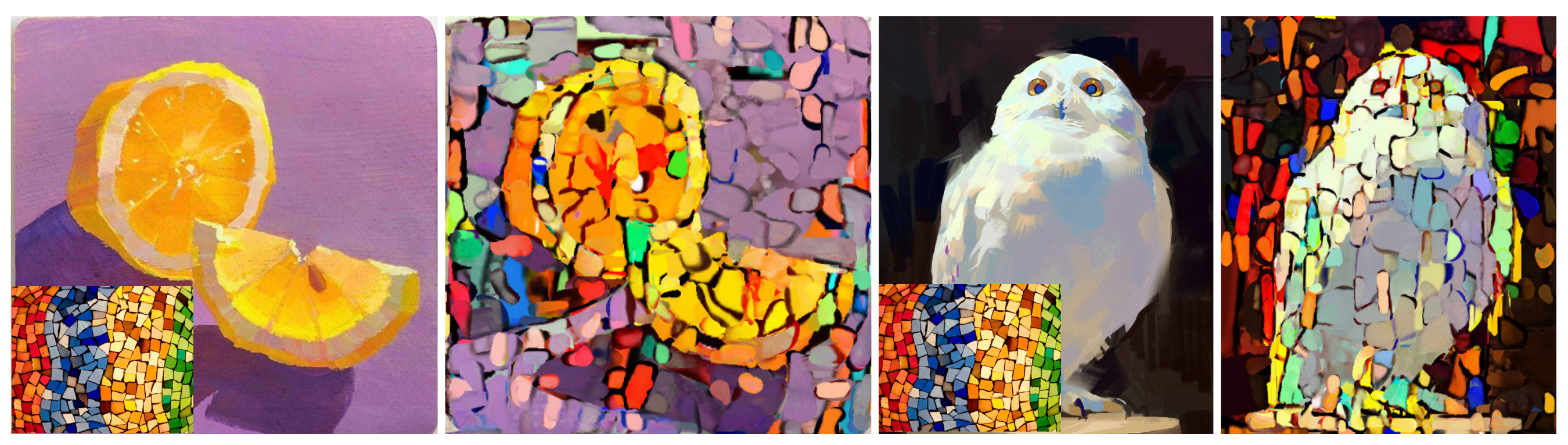}
    \vspace{-2mm}
    \caption{\textbf{Style Transfer.} Using an input image and a style guidance image (bottom left), the proposed method achieves visually consistent and high-fidelity style transfer, preserving both content structure and artistic characteristics.
    }
    \label{fig:style}
    \vspace{-2mm}
\end{figure}

\paragraph{Style Transfer.}
Our unified geometry–texture stroke representation can be naturally decoupled to enable style transfer. To achieve this, we first perform paint and smudge stroke reconstruction across all grids, rather than alternating between them, to extract their geometric structures from the input image. During the style painting phase, we modify the perceptual loss to be computed with respect to a style reference image, while keeping all other loss terms defined by the input image. This design allows the system to preserve the structural layout and semantics of the input painting while adapting its color distribution, brush texture, and material characteristics to match the target artistic style. As shown in Fig.~\ref{fig:style}, our method effectively transfers the desired artistic style while maintaining geometric structure.


\begin{figure*}[t]
    \centering
    \includegraphics[width=0.83\textwidth]{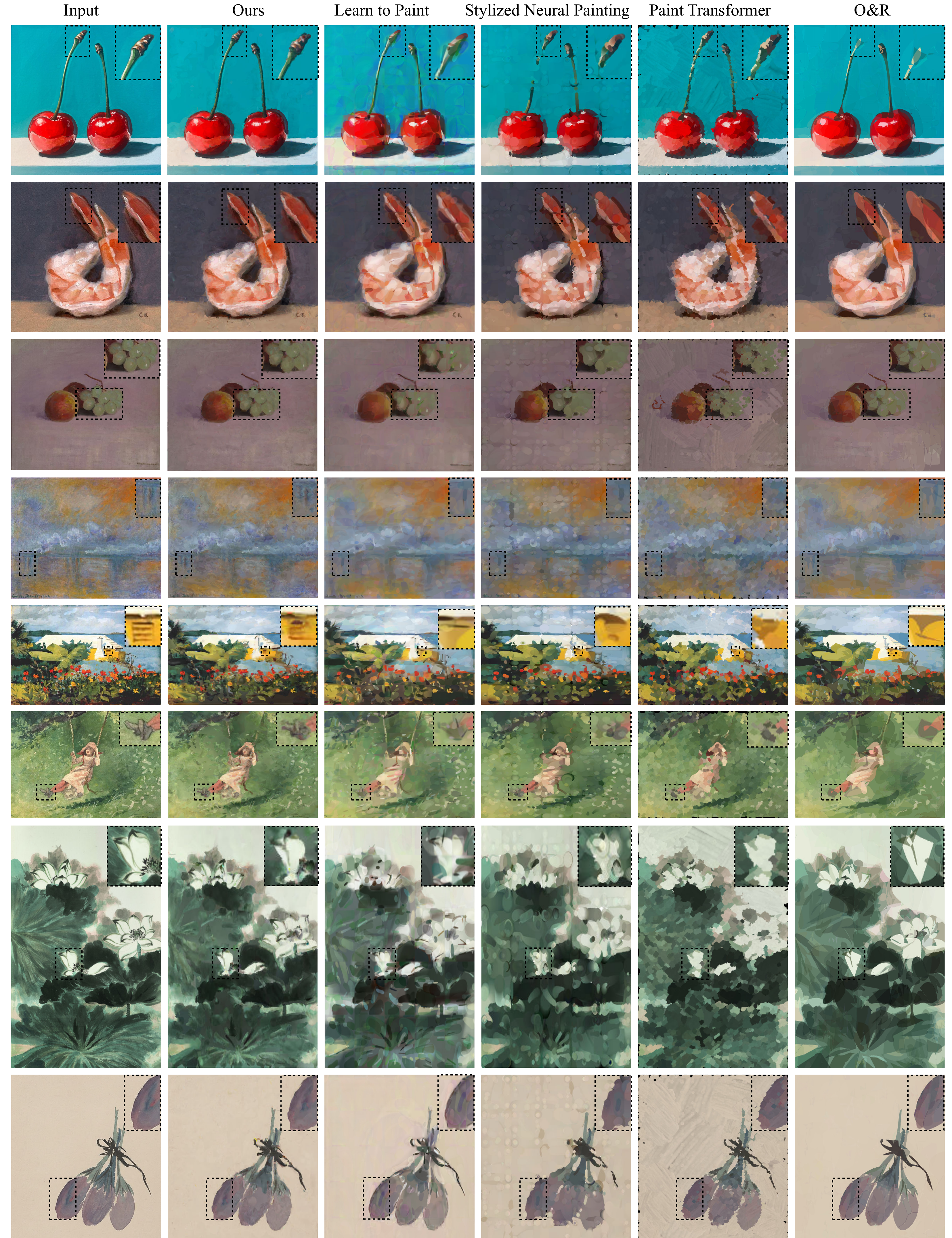}
    \vspace{-5pt}
    \caption{
        \textbf{Qualitative Comparison.} We compare our results with Stylized Neural Painting~\cite{zou2021stylized}, Paint Transformer~\cite{liu2021paint}, Learning to Paint~\cite{huang2019learning}, and O\&R~\cite{hirschorn2024optimize}. Our method produces more coherent shading and faithful texture details while preserving an aesthetically pleasing painterly style. 
    }
    \label{fig:qua}
    \vspace{-5pt}
\end{figure*}

\section{Evaluation \& Comparison}
\label{sec:eval}

\subsection{Datasets and Baselines}
\label{sec:eval_bench}
\paragraph{Datasets}
We comprehensively evaluate our method across diverse painting domains, including watercolor, oil, ink, and digital paintings. For analog watercolor, oil, and ink paintings, we use the publicly available WikiArt dataset~\cite{mao2017deepart}, from which we extract 20 paintings per category. For digital paintings, we select 20 digitally recreated artworks from an online painting website\footnote{\url{https://www.pinterest.com/}}, serving as representative examples of modern digital art styles. In total, we evaluate our method on 80 paintings to assess its effectiveness and expressiveness across a wide range of artistic styles.

\begin{figure*}[t]
\centering
\includegraphics[width=0.83\textwidth]{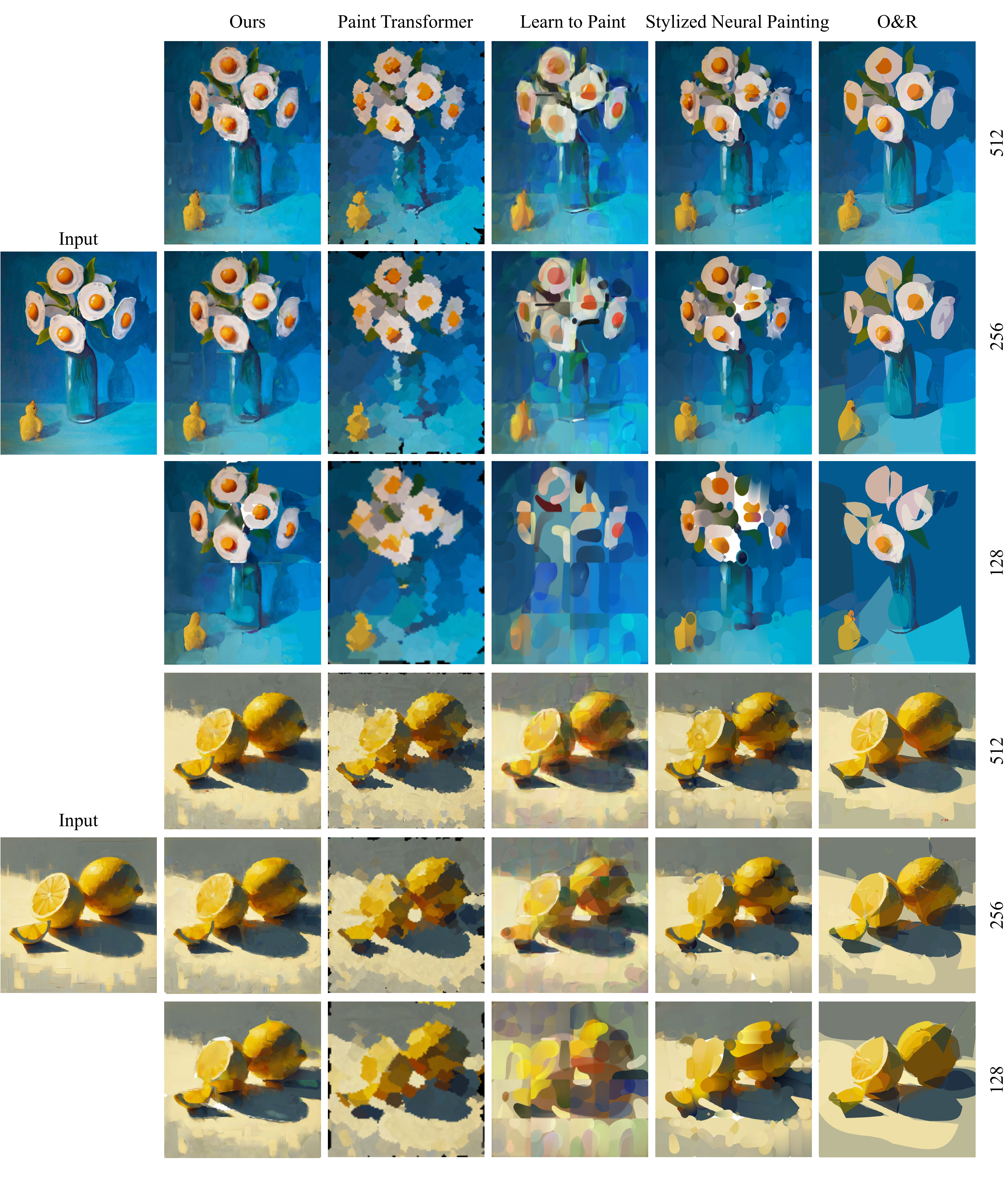}
\vspace{-10pt}
\caption{\textbf{Qualitative comparison on 128, 256 and 512 strokes.} Compared to the baseline methods, our approach more effectively captures geometric contours and produces natural shading effects, even with sparse stroke numbers. }
\label{fig:qua_many}
\vspace{-10pt}
\end{figure*}

\paragraph{Baselines}
We compare our method with three baselines that model the painting process using open strokes as primitives, similar to our framework: Stylized Neural Painting~\cite{zou2021stylized}, Paint Transformer~\cite{liu2021paint}, and Learning to Paint~\cite{huang2019learning}. Additionally, we compare with O\&R~\cite{hirschorn2024optimize}, which uses closed regions as primitives to synthesize painterly renderings. For all baselines, we adopt a $4 \times 4$ canvas partition (when applicable) assess reconstruction fidelity across varying stroke densities.

\begin{table*}[t]
\centering
\footnotesize
\renewcommand{\arraystretch}{1.2}
\setlength{\tabcolsep}{4pt}
\caption{\textbf{Quantitative Comparison.} We evaluate overall average PSNR, SSIM, LPIPS, and FD across all painting styles. Higher PSNR/SSIM and lower LPIPS/FD indicate better reconstruction fidelity. 
}
\begin{tabular}{l|cccc|cccc|cccc}
\hline
\multirow{2}{*}{\textbf{Package}} &
\multicolumn{4}{c|}{\textbf{128 Brushes}} &
\multicolumn{4}{c|}{\textbf{256 Brushes}} &
\multicolumn{4}{c}{\textbf{512 Brushes}} \\
\cline{2-13}
 & \textbf{PSNR}$\uparrow$ & \textbf{SSIM}$\uparrow$ &\textbf{LPIPS}$\downarrow$ & \textbf{FD}$\downarrow$
 & \textbf{PSNR}$\uparrow$ & \textbf{SSIM}$\uparrow$ & \textbf{LPIPS}$\downarrow$ & \textbf{FD}$\downarrow$
 & \textbf{PSNR}$\uparrow$ & \textbf{SSIM}$\uparrow$ & \textbf{LPIPS}$\downarrow$ & \textbf{FD}$\downarrow$ \\
\hline
Optimize\&Reduce
 & 19.108 & \tecell{0.5855} & \tecell{0.5637} & 13.91
 & 20.652 & \tecell{0.6032} & \tecell{0.5174} & 13.03
 & 22.429 & \tecell{0.6243} & \tecell{0.4700} & 11.53 \\
Paint Transformer
 & 16.506 & 0.5324 & 0.6775 & \tecell{13.73}
 & 17.211 & 0.5380 & 0.6285 & \tecell{12.78}
 & 18.584 & 0.5521 & 0.5560 & 11.78 \\
Learning to Paint
 & 19.181 & 0.5544 & 0.6309 & 16.84
 & \tecell{21.521} & 0.5847 & 0.5700 & 14.82
 & \tecell{23.155} & 0.6145 & 0.5231 & \tecell{11.27} \\
Stylized-Neural-Painting
 & \tecell{20.288} & 0.5721 & 0.5654 & 16.45
 & 21.154 & 0.5896 & 0.5272 & 15.53
 & 21.430 & 0.6029 & 0.5006 & 13.70 \\

\textbf{Ours}
 & \apcell{21.463} & \apcell{0.5993} & \apcell{0.5064} & \apcell{9.75}
 & \apcell{22.444} & \apcell{0.6168} & \apcell{0.4711} & \apcell{8.18}
 & \apcell{23.241}& \apcell{0.6335}& \apcell{0.4362} & \apcell{6.64} \\ 
\hline
\end{tabular}
\label{tab:overall}
\end{table*}

\subsection{Quantitative Comparison}
\label{sec: quantitative eval}



To quantitatively assess reconstruction quality, we compare our generated results with the baselines on the batch of 80 paintings, using configurations of 128, 256, and 512 strokes or regions, resulting in a total of 240 paintings for each method.
We employ multiple complementary metrics, including Peak Signal-to-Noise Ratio (PSNR)~\cite{gonzalez2009digital}, Structural Similarity Index (SSIM)~\cite{wang2004image}, Learned Perceptual Image Patch Similarity (LPIPS)~\cite{zhang2018unreasonable}, and ResNet-based feature distance (FD)~\cite{he2016deep}. PSNR and SSIM evaluate pixel-level fidelity and structural consistency between the reconstructed and reference paintings, reflecting low-level reconstruction accuracy. LPIPS measures perceptual similarity in a latent space, providing a more reliable indicator of texture coherence and structural realism. FD assesses the distributional difference between generated and reference images in the feature space of an Inception network, serving as a high-level perceptual metric for overall realism and style consistency.


The quantitative evaluation results (Table~\ref{tab:overall}) demonstrate that our method consistently outperforms all baselines across different metrics. Notably, under sparse stroke settings (128 strokes), our approach achieves a significant improvement over the others, indicating that the proposed framework can reconstruct images effectively even with limited strokes, thanks to its more expressive stroke representation. While O\&R achieves the second-best performance at 128 and 256 strokes, its improvement diminishes at 512 strokes. Although at higher stroke counts the PSNR of our method is comparable to that of Learning to Paint, our LPIPS and FD scores remain substantially better, suggesting that while those methods may recover coarse color alignment, they struggle to produce perceptually coherent strokes. Overall, our method achieves superior reconstruction fidelity and perceptual realism across varying stroke densities. Detailed per-category metric analyses are provided in the supplementary document.

\subsection{Qualitative Comparison}
We present qualitative comparisons with baseline methods in Fig.~\ref{fig:qua} and Fig.~\ref{fig:qua_many}. In Fig.~\ref{fig:qua} we showcase the paintings generated using 1024-stroke reconstructions arranged in a $5 \times 5$ grid across diverse subjects and painting styles. Our method produces reconstructions that are both faithful to the target content and rich in painterly texture. 
Additionally, we visualize the reconstruction results across different sparse stroke settings (128, 256, 512 strokes) in Fig.~\ref{fig:qua_many}.
We observe that with sufficient stroke budgets (Fig.~\ref{fig:qua}), Learning to Paint \cite{huang2019learning} often produces coarse, grid-like patterns and oversimplified regions, especially in high-frequency areas such as floral petals and grape clusters. PaintTransformer \cite{liu2021paint}, while exploring predefined brush shapes, tends to generate boundary leaks and patchy artifacts and remains largely constrained to oil-like appearances due to its reliance on a single brush type. Stylized Neural Painting \cite{zou2021stylized} achieves visually appealing global tones but frequently loses local detail. O\&R \cite{hirschorn2024optimize} accurately reproduces global color distributions; however, its closed-region primitives make it difficult to reproduce in real digital painting software, as irregular enclosed shapes are challenging to replicate with standard brush operations. Under sparse stroke settings (Fig.~\ref{fig:qua_many}), both our method and O\&R capture the overall structure effectively with only 128 strokes, while LearningToPaint, Stylized Neural Painting, and PaintTransformer struggle to preserve contours and object boundaries. This demonstrates that our approach can better capture geometric and structural information even under limited stroke numbers. Moreover, almost all baseline methods lack color blending, resulting in hard shading effects. In contrast, our framework reconstructs natural shading and realistic tonal variations. Across all subjects and styles, our method achieves a robust balance between global structure and fine detail, producing results that are visually closer to the targets and more naturally painterly.

\subsection{Ablation Study}

We conduct ablation studies on key components of our framework, using the same input paintings as in Section~\ref{sec:eval_bench}. These experiments evaluate the contribution of each proposed component to the final painting reconstruction quality, as well as the effect of varying stroke numbers on the overall reconstruction performance.

\begin{figure}[t]
    \centering
    \includegraphics[width=\columnwidth]{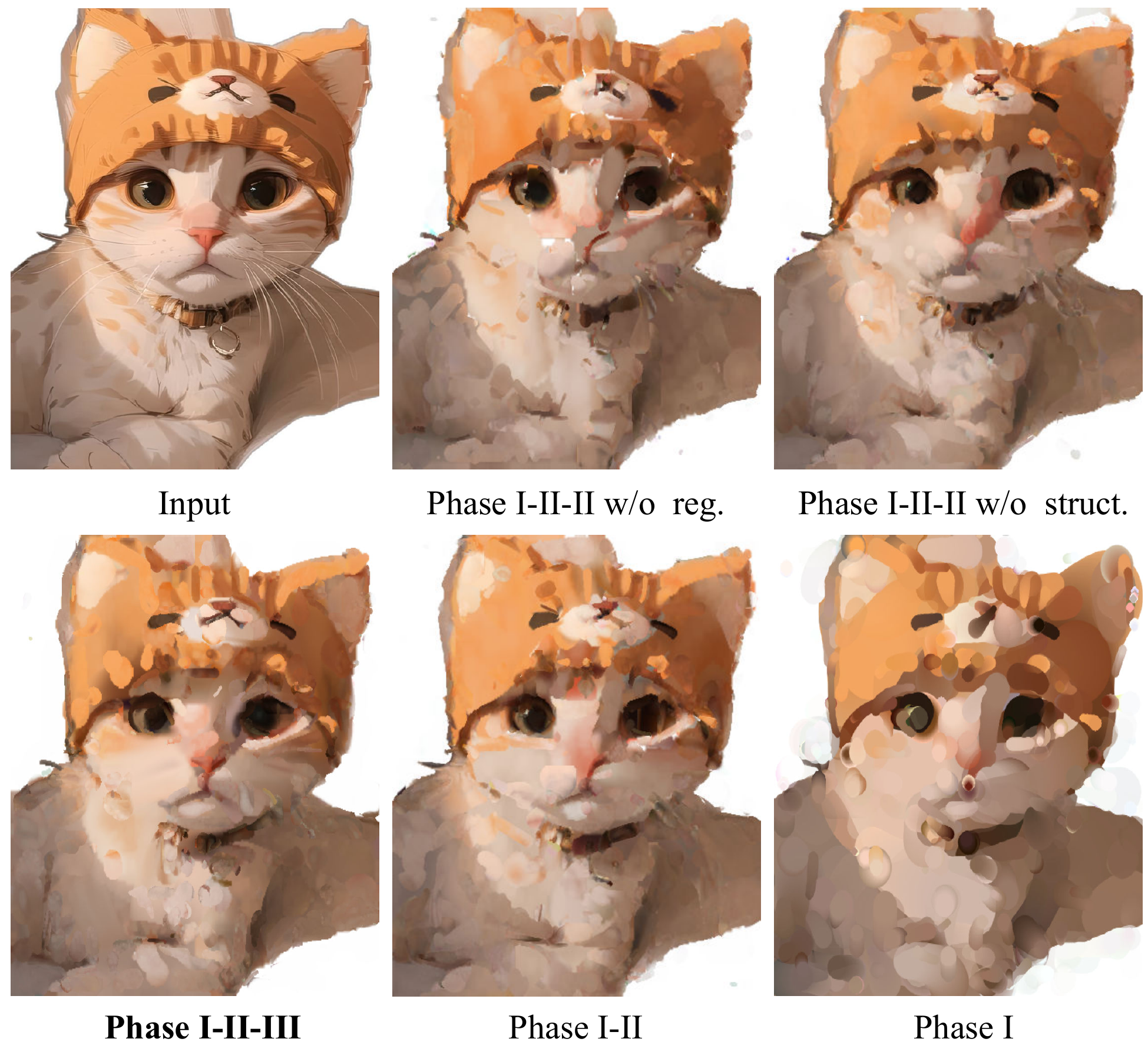}
    \vspace{-2mm}
    \caption{\textbf{Ablation Study.} 
    We present results from different training stages: Phase I only, Phases I–II, and Phases I–II–III (full pipeline), as well as results obtained without regularization guidance and without structural guidance.
    We observe that the reconstruction quality consistently improves across all training phases, producing more accurate and coherent painting results, while both regularization and structural guidance play a crucial role in ensuring reconstruction fidelity.}

    \label{fig:ablation}
    \vspace{-2mm}
\end{figure}

\begin{figure*}[h]
    \centering
    \includegraphics[width=0.8\textwidth]{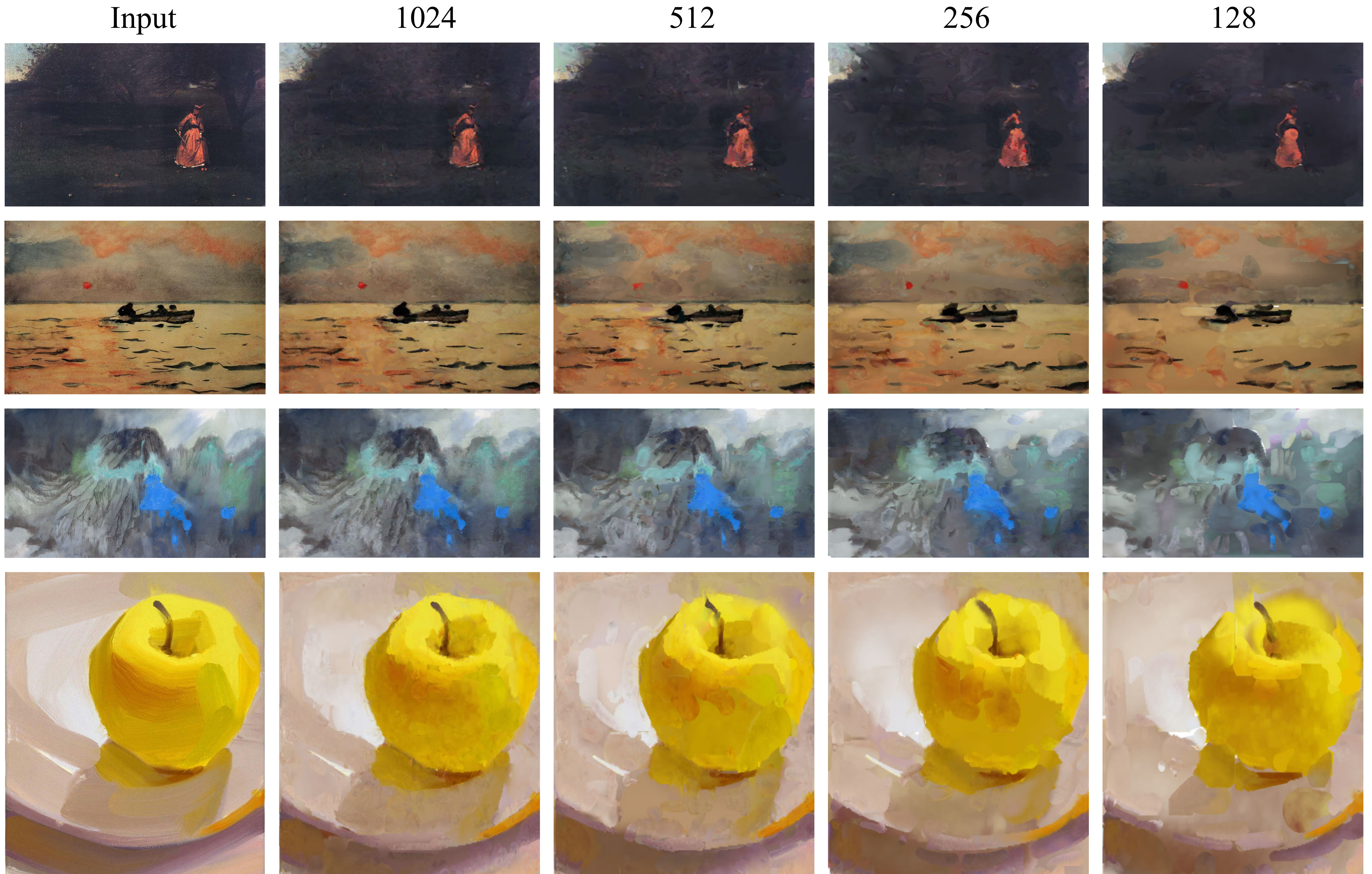}
    \caption{\textbf{Stroke Number Ablation}
       From left to right: the input image followed by reconstructed results using 1024, 512, 256, and 128 strokes, respectively. The results demonstrate that the proposed method accurately reconstructs paintings across varying stroke counts and diverse painting styles.
    }
    \label{fig:strokenum}
\end{figure*}

\begin{table}[t]
\centering
\footnotesize
\caption{\textbf{Quantitative Ablation Study.} We evaluate the results on 512 brushstrokes across different stages and components of our method using the same metric as above.}
\label{tab:ablation}
\begin{tabular}{l|cccc}
\hline
\textbf{Method} & \textbf{PSNR} $\uparrow$ & \textbf{SSIM} $\uparrow$ & \textbf{LPIPS} $\downarrow$ & \textbf{FD} $\downarrow$ \\
\hline
Phase~I & 22.019 & 0.6695 & 0.4731 & 11.94  \\
Phase~I-II & 23.699 & 0.6880 & 0.4324 & 5.57\\
Phase~I-II-III w/o Structure Guidance & 23.426 & 0.6853 & 0.4405 & 5.90   \\
Phase~I-II-III w/o Regularization & 22.750 & 0.6765 & 0.4629 & 6.64   \\
\hline
\textbf{Phase~I-II-III (Ours)} & \apcell{23.898} & \apcell{0.6909} & \apcell{0.4248} & \apcell{5.46}   \\
\hline
\end{tabular}
\end{table}

\paragraph{Reconstruction Framework Design.}
We first evaluate the effectiveness of our three-phase framework design, which progressively refines painting reconstruction through paint stroke appearance, texture stylization, and smudge stroke optimization. As shown in Table~\ref{tab:ablation}, advancing from Phase~I to Phase~III consistently improves reconstruction quality. Phase~I, which focuses solely on paint stroke appearance reconstruction with dual-color strokes, captures the overall structure but lacks stylized texture and smooth shading effects. Incorporating Phase~II for texture stylization (Phase~I–II) significantly enhances color fidelity and artistic expressiveness by generating diverse stroke textures that better match various painting styles. Extending to Phase~III with smudge stroke reconstruction (Phase~I–II–III) further improves shading continuity and tonal smoothness, producing more natural and perceptually coherent renderings. As illustrated in Fig.~\ref{fig:ablation}, the cat’s face exhibits softer and more natural shading transitions, instead of the harsh, segmented appearance caused by discrete strokes.

\paragraph{Structural Guidance.} Two structure-guided losses, the gradient-based alignment loss $\mathcal{L}_{\text{grad}}$ and the layer-based segmentation loss $\mathcal{L}_{\text{seg}}$, are adopted to ensure that reconstructed strokes accurately follow the intrinsic geometry of the painting. The segmentation loss $\mathcal{L}_{\text{seg}}$ constrains strokes to stay within object boundaries, thereby avoiding unwanted strokes across adjacent objects. The $\mathcal{L}_{\text{grad}}$, composed of orientation and magnitude terms, enforces the stroke flow to align with local edges and contours. As illustrated in Fig.~\ref{fig:ablation}, the absence of $\mathcal{L}_{\text{grad}}$ and $\mathcal{L}_{\text{seg}}$ leads to irregular boundaries near the cat’s right ear and disordered strokes on the forehead, introducing structural ambiguity and visual inconsistency. Together, these two losses contribute to geometrically aligned, semantically consistent, and perceptually coherent stroke reconstructions.

\paragraph{Regularization}
To avoid vanishing gradients when strokes are far from their target regions and to prevent them from collapsing into excessively small or vanishing areas, we introduce the optimal transport loss $\mathcal{L}_{\text{OT}}$ and the area regularization loss $\mathcal{L}_{\text{area}}$. As shown in Fig.~\ref{fig:ablation}, removing $\mathcal{L}_{\text{OT}}$ and $\mathcal{L}_{\text{area}}$ leads to fragmented or missing strokes around the cat’s nose and hat, as well as noticeable color mismatches in the mouse region, resulting in sparse, unstable, and perceptually inconsistent reconstructions.

\paragraph{Stroke Number.} We assess the influence of stroke count on reconstruction quality. As illustrated in Fig.~\ref{fig:strokenum}, increasing the stroke count (128, 256, 512, and 1024) across four representative styles including watercolor, oil, digital, and ink progressively improves the reconstruction fidelity, resulting in finer structural details and richer tonal variations. Moreover, even with a low stroke count (128 or 256), our method effectively preserves the global structure and maintains perceptually consistent shading, demonstrating strong robustness and efficient stroke utilization in capturing complex visual content.

\subsection{User Study}

To evaluate the perceptual quality and human preference of generated results, we conduct a user study involving fourteen participants. The study follows a perceptual evaluation protocol comparing our generative results against several state-of-the-art painting generation methods using a two-alternative forced choice (2AFC) design \cite{gal2024breathing, xie2025physanimator}.

\paragraph{Participants and Procedure.}
A total of 14 participants (including $11$ novice users and $3$ professional artists) are recruited for the study. During the experiment, participants are presented with pairs of generative painting results—one produced by our method and the baseline methods.
The order of presentation is randomized to minimize bias. Following the 2AFC protocol, participants were asked to choose their preferred reconstructed paintings in each pair according to five metrics: (a) Color Alignment (CA) assesses color consistency in hue, tone, and saturation;  (b) Texture and Shading (TS) measures the realism of surface texture and light–shadow transitions; (c) Painting Style (PS) evaluates the coherence of strokes, palette, and abstraction with the intended style; (d) Shape Fidelity (SF) reflects the structural accuracy of objects and brushstrokes; and (e) Overall Quality (OQ) represents the overall visual realism and aesthetic appeal.

\begin{table}[h]
\centering
\caption{\textbf{User Study.} We evaluate 2AFC user preferences against the baselines across five evaluation metrics: Color Alignment (CA), Painting Style (PS), Texture and Shading (TS), Shape Fidelity (SF), and Overall Quality (OQ). A score above 50\% indicates that participants prefer our results over the respective baseline.}
\setlength\tabcolsep{8.0pt}
\label{tab:user_study}
\footnotesize{
\begin{tabular}{p{1.1in}|ccccc}
\hline
\textbf{Method} & $\textbf{CA}\uparrow$ & $\textbf{TS}\uparrow$ & $\textbf{PS}\uparrow$ & $\textbf{SC}\uparrow$ & $\textbf{OQ}\uparrow$ \\ \hline
Stylized Neural Painting & 75\% & 78\% & 78\% & 78\% & 77\% \\
Paint Transformer & 82\% & 85\% & 81\% & 85\% & 84\% \\
Learning to Paint & 78\% &77\% & 76\% & 78\% & 77\% \\
O\&R & 66\% & 73\% & 72\% & 76\% & 70\% \\ \hline
\end{tabular}
}
\end{table}

In Table~\ref{tab:user_study}, we report the itemized user preferences for each baseline (whether users prefer our results over the baseline; a score above 50\% indicates an overall more preferred result). Our method is consistently preferred across all evaluation criteria and baseline methods. Compared with the open-curve–based painting baselines (Stylized Neural Painting, Paint Transformer, and Learning to Paint), which tend to produce visually similar painterly textures and styles, and O\&R (a closed-region–based vector graphics method) offers stronger color alignment. However, our method achieves superior performance across all perceptual dimensions, excelling not only in color alignment but also in producing more realistic textures and expressive styles. Participants particularly emphasized the natural color harmony, smooth shading effects, and diverse painting styles, ranging from oil to ink, in our generated results, highlighting the improved perceptual realism and artistic expressiveness achieved by the proposed framework.

\section{Conclusions}
\label{sec:con}
We propose a differentiable stroke reconstruction framework that novelly unifies stroke-based paint and smudge rendering with styled texture generation. Our method reproduces the iterative painting–smudging loop, yielding realistic painterly effects with smooth shading and expressive strokes. The framework generates stroke-based painting processes that emulate human-like painting and smudging, capturing the dynamic evolution of artwork while maintaining both structural fidelity and aesthetic expressiveness. We introduce a unified stroke representation that jointly models paint and smudge strokes, enabling consistent simulation of stroke deposition and color blending. A stylized texture reconstruction further enriches the results by producing diverse and visually coherent textures across different painting styles. Versatile experiments show our method generates realistic and expressive painting processes. 

\paragraph{Limitation and Future Work}
Our approach has several limitations that suggest directions for future work. First, the current patch-based optimization strategy for painting and smudging, while effective, remains somewhat rigid. The order of painting and smudging operations within each grid patch is predefined, limiting flexibility. Enabling dynamic alternation between painting and smudging strokes could better emulate the adaptive, iterative nature of human painting workflows.

Another limitation lies in the stroke representation. Currently, stroke shapes are restricted to open Bézier curve parameterizations, which may not fully capture the irregular and diffusive boundaries often observed in watercolor or ink paintings. Incorporating more expressive stroke models, such as displacement-map–based deformations or parameterized boundary perturbations, could enable the reproduction of complex zigzag and diffusion patterns characteristic of fluid-based media.

\bibliographystyle{ACM-Reference-Format}
\bibliography{reference}

\clearpage
\appendix
\onecolumn   

\section{Per-Category Quantitative Results}
\label{app:subcategory}

We further provide per-category quantitative evaluations in Table~\ref{tab:allcategories-vertical}, covering four representative painting styles: watercolor, oil, ink, and digital.
Across all metrics (PSNR, SSIM, LPIPS, and FD) and stroke budgets (128, 256, and 512), our method consistently achieves superior performance compared to existing approaches, demonstrating both higher reconstruction fidelity and better perceptual quality.

\begin{table*}[h]
\centering
\footnotesize
\renewcommand{\arraystretch}{1.2}
\setlength{\tabcolsep}{4pt}
\caption{Per-category quantitative results: PSNR, SSIM, LPIPS, and FD for 128, 256, and 512 brush strokes.}
\begin{tabular}{l|cccc|cccc|cccc}
\hline
\multicolumn{13}{c}{\textbf{Watercolor}}\\
\hline
\multirow{2}{*}{\textbf{Method}} &
\multicolumn{4}{c|}{128 Brushes} &
\multicolumn{4}{c|}{256 Brushes} &
\multicolumn{4}{c}{512 Brushes}\\
\cline{2-13}
 & PSNR$\uparrow$ & SSIM$\uparrow$ & LPIPS$\downarrow$ & FD$\downarrow$
 & PSNR$\uparrow$ & SSIM$\uparrow$ & LPIPS$\downarrow$ & FD$\downarrow$
 & PSNR$\uparrow$ & SSIM$\uparrow$ & LPIPS$\downarrow$ & FD$\downarrow$\\
\hline
Optimize\&Reduce&
18.872 & 0.5327 & 0.6093 & 15.30 &
20.301 & 0.5526 & 0.5590 & 14.51 &
21.782 & 0.5759 & 0.5058 & 12.16 \\
PaintTransformer&
16.594 & 0.4915 & 0.6854 & 13.33 &
17.175 & 0.4956 & 0.6276 & 12.28 &
18.548 & 0.5093 & 0.5457 & 11.79 \\
LearningToPaint&
19.459 & 0.5183 & 0.6475 & 17.04 &
21.592 & 0.5534 & 0.5800 & 15.48 &
23.156 & 0.5884 & 0.5294 & 11.86 \\
Stylized-Neural-Painting&
20.269 & 0.5310 & 0.5880 & 17.12 &
20.994 & 0.5521 & 0.5451 & 16.11 &
21.374 & 0.5705 & 0.5122 & 14.21 \\
\textbf{Ours}& 
21.168 & 0.5655 & 0.5188 & 10.13 &
22.290 & 0.5855 & 0.4905 & 9.05 &
23.269 & 0.6085 & 0.4499 & 7.00 \\
\hline\hline

\multicolumn{13}{c}{\textbf{OilPainting}}\\
\hline
\multirow{2}{*}{\textbf{Method}} &
\multicolumn{4}{c|}{128 Brushes} &
\multicolumn{4}{c|}{256 Brushes} &
\multicolumn{4}{c}{512 Brushes}\\
\cline{2-13}
 & PSNR$\uparrow$ & SSIM$\uparrow$ & LPIPS$\downarrow$ & FD$\downarrow$
 & PSNR$\uparrow$ & SSIM$\uparrow$ & LPIPS$\downarrow$ & FD$\downarrow$
 & PSNR$\uparrow$ & SSIM$\uparrow$ & LPIPS$\downarrow$ & FD$\downarrow$\\
\hline
Optimize\&Reduce&
19.791 & 0.4875 & 0.6395 &14.26 &
21.320 & 0.5103 & 0.5903 & 13.96 &
22.789 & 0.5342 & 0.5345 & 12.98 \\
PaintTransformer&
17.430 & 0.4471 & 0.7152 & 12.54 &
17.937 & 0.4513 & 0.6673 & 11.88 &
19.420 & 0.4712 & 0.5917 & 11.46 \\
LearningToPaint&
20.695 & 0.4796 & 0.6746 & 16.63 &
22.785 & 0.5178 & 0.6080 & 15.62 &
24.201 & 0.5537 & 0.5586 & 11.93 \\
Stylized-Neural-Painting&
21.226 & 0.4913 & 0.6246 & 16.89 &
22.062 & 0.5163 & 0.5798 & 16.49 &
22.228 & 0.5359 & 0.5459 & 14.96 \\
\textbf{Ours}&
22.439 & 0.5268 & 0.5508 & 10.79 &
23.318 & 0.5494 & 0.5154 & 9.26 &
24.026 & 0.5709 & 0.4743 & 7.16 \\
\hline\hline

\multicolumn{13}{c}{\textbf{InkPainting}}\\
\hline
\multirow{2}{*}{\textbf{Method}} &
\multicolumn{4}{c|}{128 Brushes} &
\multicolumn{4}{c|}{256 Brushes} &
\multicolumn{4}{c}{512 Brushes}\\
\cline{2-13}
 & PSNR$\uparrow$ & SSIM$\uparrow$ & LPIPS$\downarrow$ & FD$\downarrow$
 & PSNR$\uparrow$ & SSIM$\uparrow$ & LPIPS$\downarrow$ & FD$\downarrow$
 & PSNR$\uparrow$ & SSIM$\uparrow$ & LPIPS$\downarrow$ & FD$\downarrow$\\
\hline
Optimize\&Reduce&
18.798 & 0.6833 & 0.5000 & 13.57 &
20.558 & 0.7009 & 0.4576 & 12.52 &
22.327 & 0.7179 & 0.4245 & 12.02 \\
PaintTransformer&
15.412 & 0.5978 & 0.6736 & 14.17 &
16.198 & 0.6036 & 0.6299 & 13.29 &
17.444 & 0.6130 & 0.5554 & 12.41 \\
LearningToPaint&
18.210 & 0.6239 & 0.6073 & 17.18 &
20.692 & 0.6493 & 0.5500 & 14.91 &
22.407 & 0.6738 & 0.5079 & 11.67 \\
Stylized-Neural-Painting&
19.287 & 0.6339 & 0.5485 & 17.07 &
20.167 & 0.6452 & 0.5189 & 16.39 &
20.490 & 0.6549 & 0.4954 & 14.62 \\
\textbf{Ours}&
20.339 & 0.6522 & 0.4962 & 9.83 &
21.309 & 0.6664 & 0.4555 & 7.94 &
22.055 & 0.6779 & 0.4236 & 6.71 \\
\hline\hline

\multicolumn{13}{c}{\textbf{DigitalPainting}}\\
\hline
\multirow{2}{*}{\textbf{Method}} &
\multicolumn{4}{c|}{128 Brushes} &
\multicolumn{4}{c|}{256 Brushes} &
\multicolumn{4}{c}{512 Brushes}\\
\cline{2-13}
 & PSNR$\uparrow$ & SSIM$\uparrow$ & LPIPS$\downarrow$ & FD$\downarrow$
 & PSNR$\uparrow$ & SSIM$\uparrow$ & LPIPS$\downarrow$ & FD$\downarrow$
 & PSNR$\uparrow$ & SSIM$\uparrow$ & LPIPS$\downarrow$ & FD$\downarrow$\\
\hline
Optimize\&Reduce&
18.972 & 0.6383 & 0.5062 & 12.53 &
20.430 & 0.6490 & 0.4629 & 11.13 &
22.816 & 0.6691 & 0.4152 & 8.95 \\
PaintTransformer&
16.587 & 0.5934 & 0.6356 & 14.88 &
17.536 & 0.6014 & 0.5890 & 13.67 &
18.925 & 0.6149 & 0.5313 & 11.48 \\
LearningToPaint&
18.359 & 0.5957 & 0.5940 & 16.52 &
21.017 & 0.6183 & 0.5417 & 13.29 &
22.857 & 0.6421 & 0.4964 & 9.61 \\
Stylized-Neural-Painting&
20.370 & 0.6323 & 0.5004 & 14.72 &
21.392 & 0.6450 & 0.4652 & 13.13 &
21.629 & 0.6501 & 0.4491 & 11.01 \\
\textbf{Ours}&
21.906 & 0.6527 & 0.4599 & 8.25 &
22.861 & 0.6660 & 0.4229 & 6.48 &
23.613 & 0.6765 & 0.3972 & 5.68 \\
\hline
\end{tabular}
\label{tab:allcategories-vertical}
\end{table*}

\newpage
\section{More Painting Results}
We showcase additional results, including 1024-stroke reconstructions arranged in a $5 \times 5$ grid and 128–512-stroke reconstructions on a $4 \times 4$ canvas, using images from the same test set. The test set includes samples from the publicly available WikiArt dataset~\cite{mao2017deepart} and recreated artworks collected from an online painting website\footnote{\url{https://www.pinterest.com/}}
, demonstrating the strong generalization ability of our method across diverse subjects and painting styles.

\begin{figure}
\centering
\includegraphics[width=0.95\textwidth]{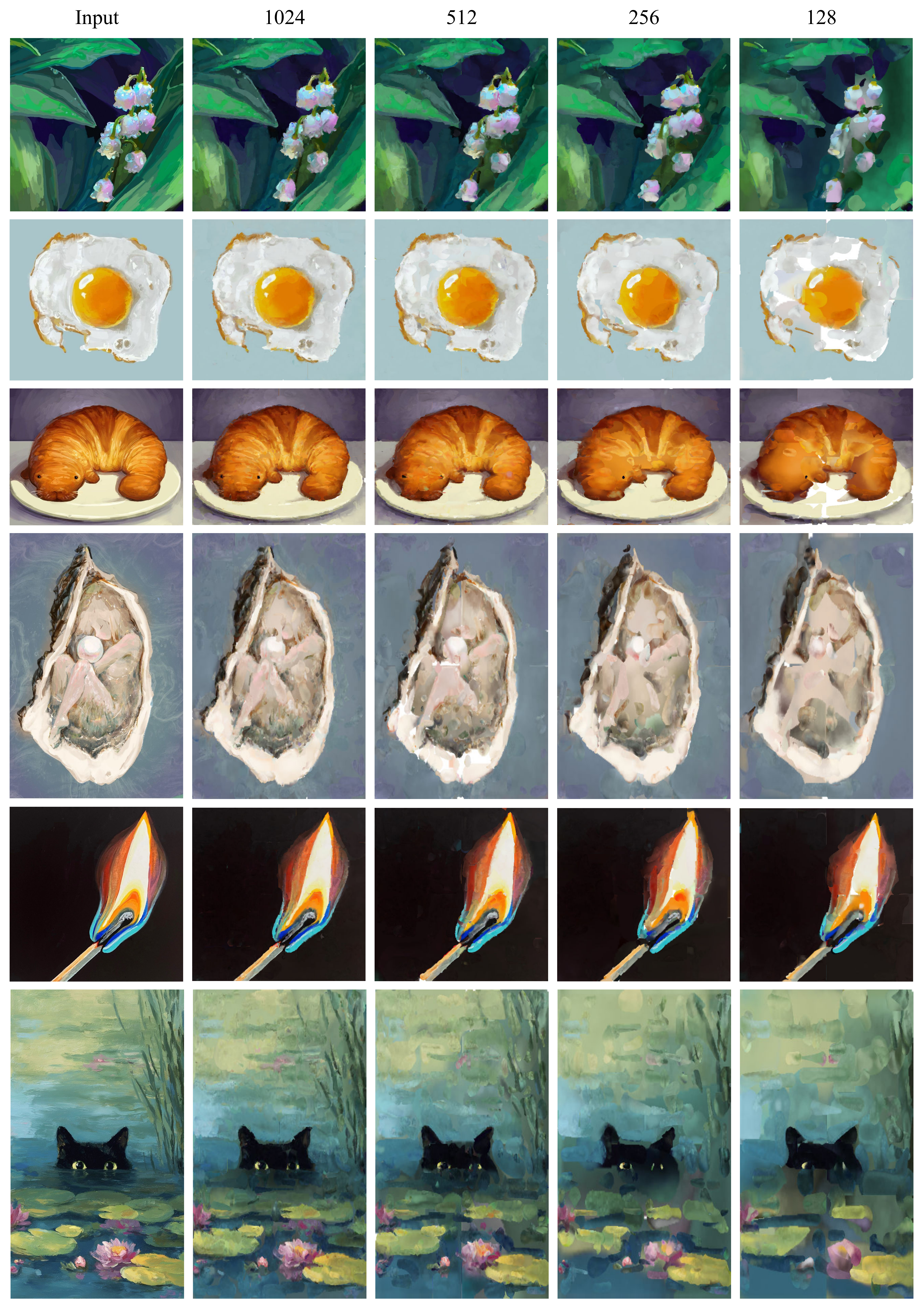}
\vspace{-10px}
\caption{More showcases from 128 to 1024 strokes in digital painting with the proposed method.}
\label{fig:digital1}
\end{figure}

\begin{figure}
\centering
\includegraphics[width=0.95\textwidth]{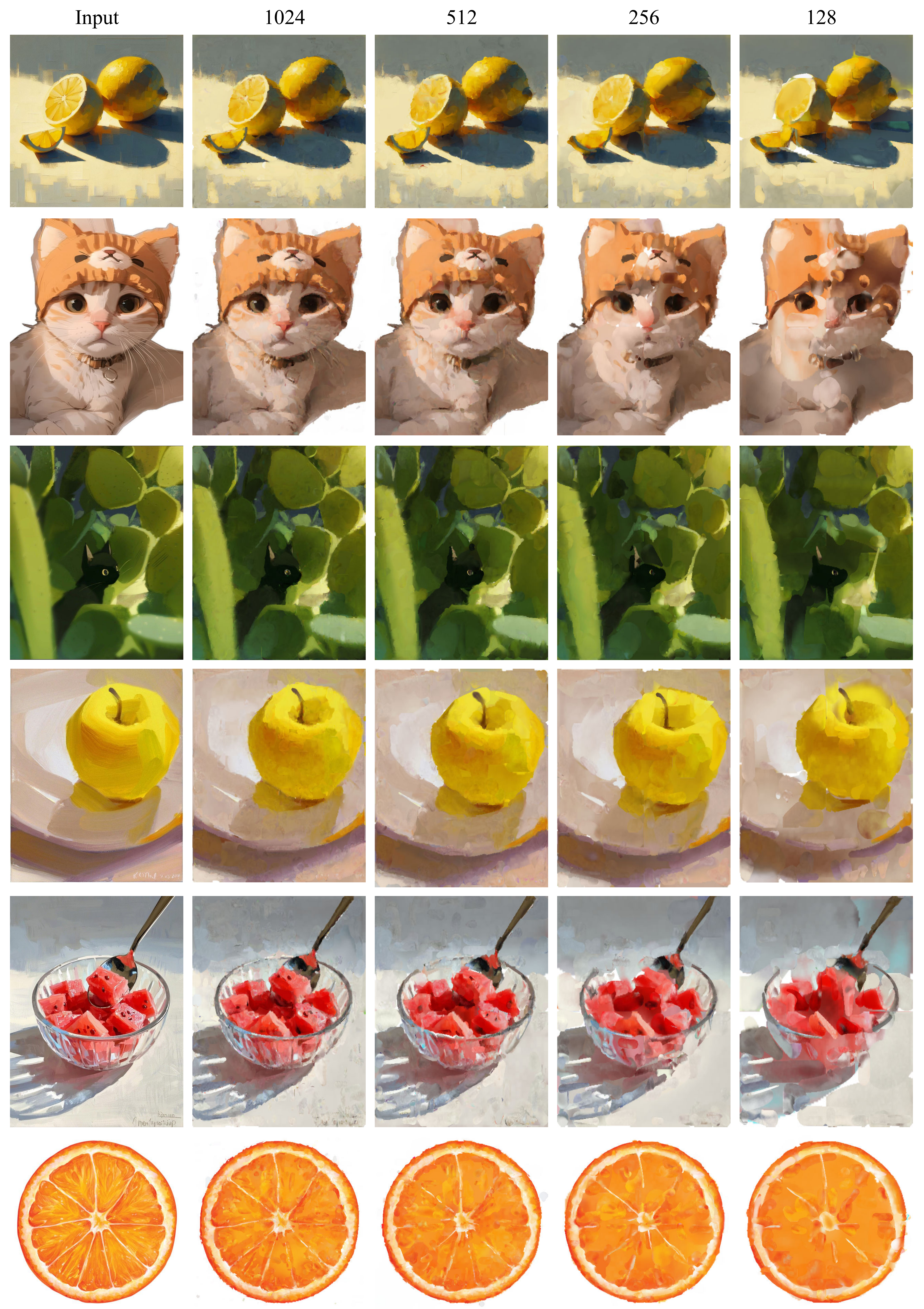}
\vspace{-10px}
\caption{More showcases from 128 to 1024 strokes in digital paintings with the proposed method.}
\label{fig:digital2}
\end{figure}

\begin{figure}
\centering
\includegraphics[width=0.95\textwidth]{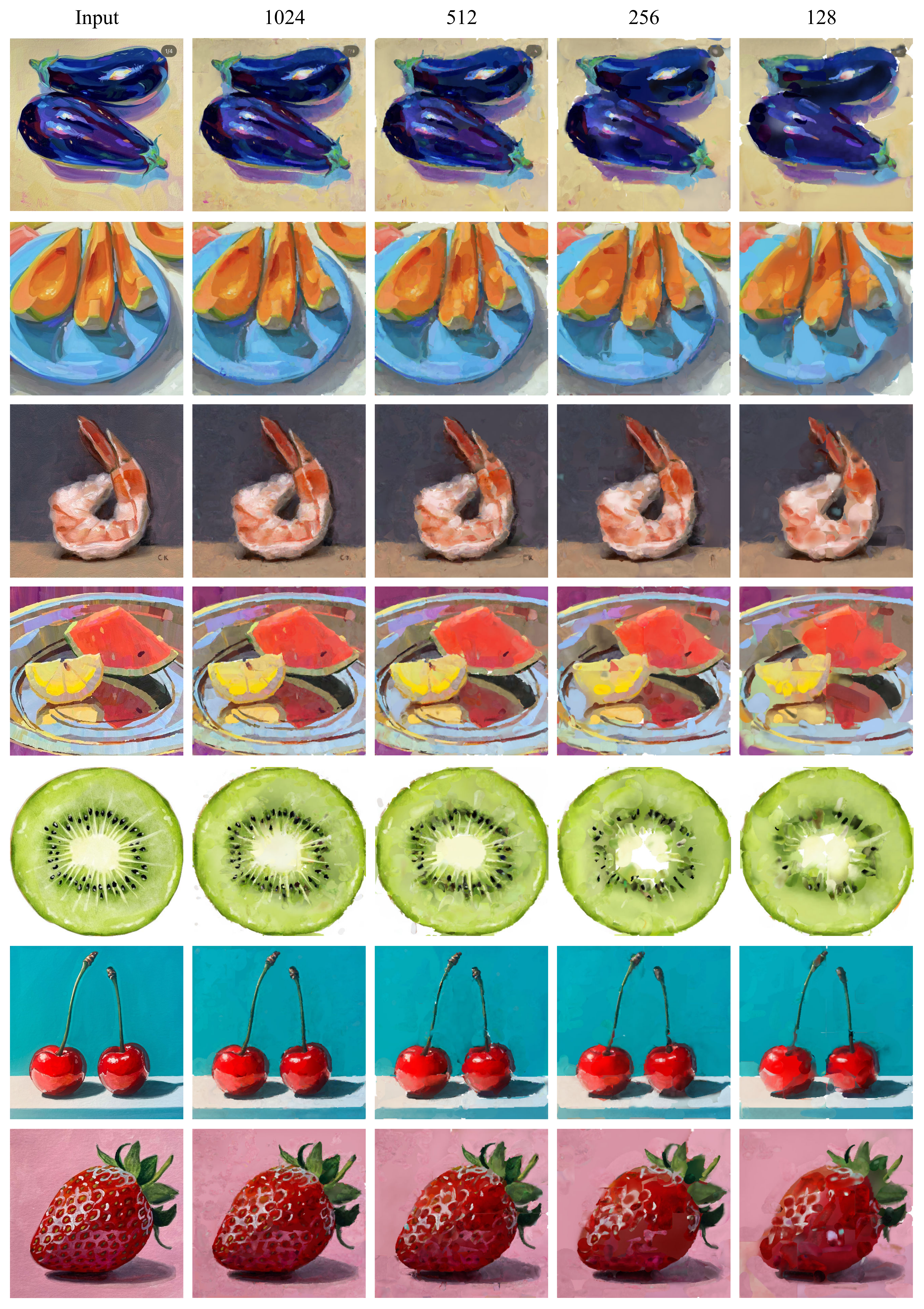}
\vspace{-10px}
\caption{More showcases from 128 to 1024 strokes in digital paintings with the proposed method.}
\label{fig:digital3}
\end{figure}

\begin{figure}
\centering
\includegraphics[width=0.95\textwidth]{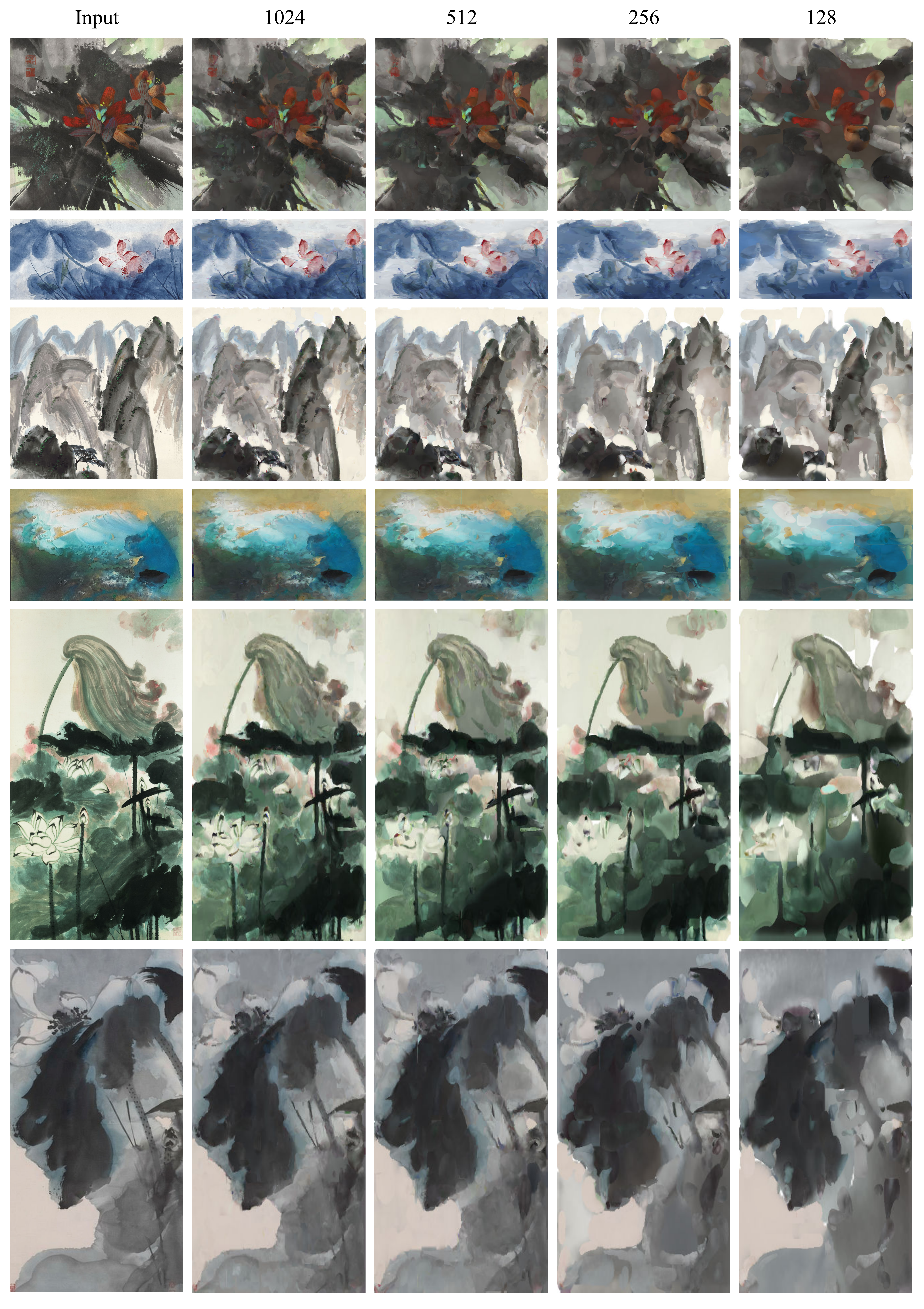}
\vspace{-15px}
\caption{More showcases from 128 to 1024 strokes in ink paintings with the proposed method.}
\label{fig:ink1}
\end{figure}

\begin{figure}
\centering
\includegraphics[width=0.95\textwidth]{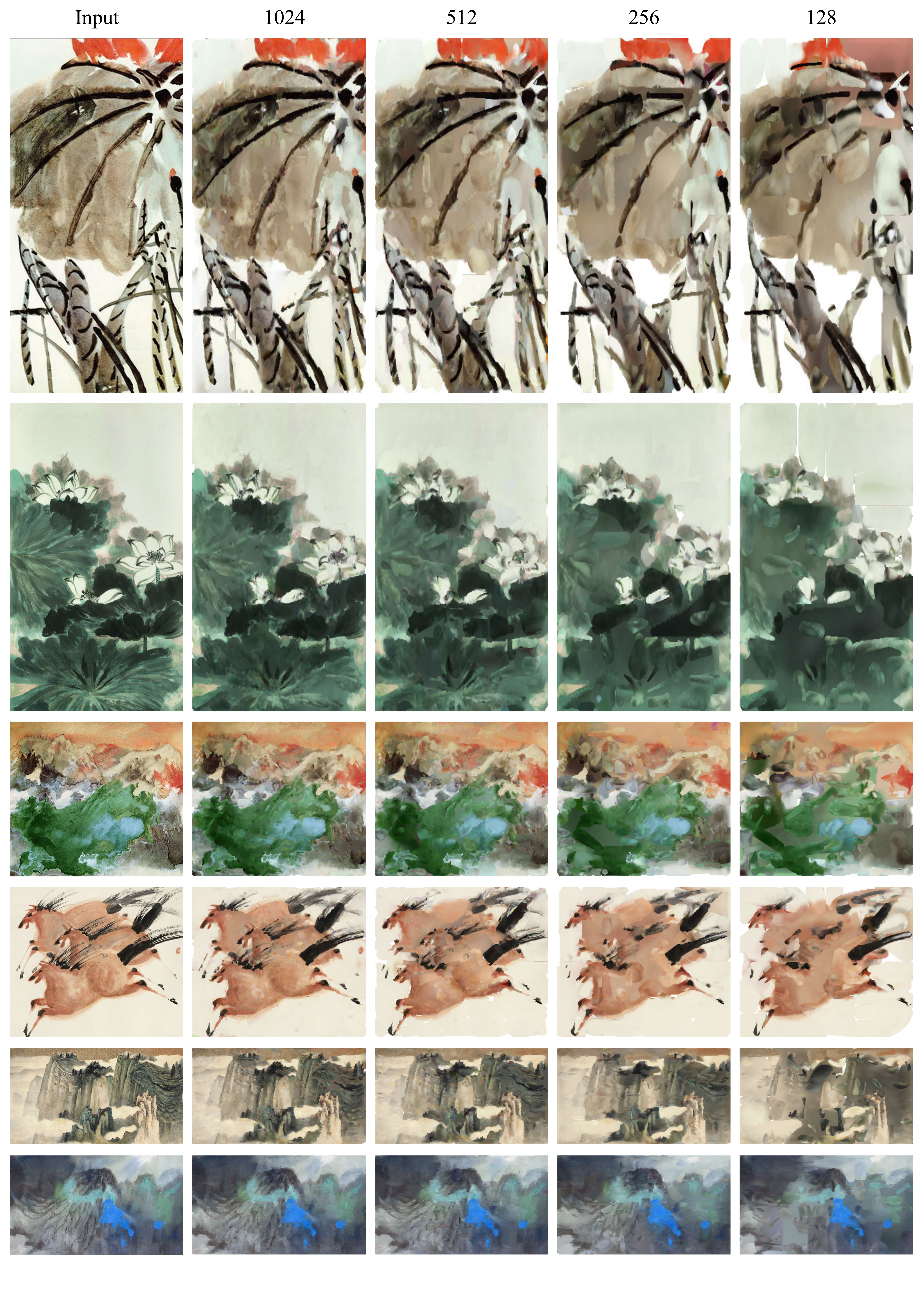}
\vspace{-30px}
\caption{More showcases from 128 to 1024 strokes in ink paintings with the proposed method.}
\label{fig:ink2}
\end{figure}

\begin{figure}
\centering
\includegraphics[width=0.95\textwidth]{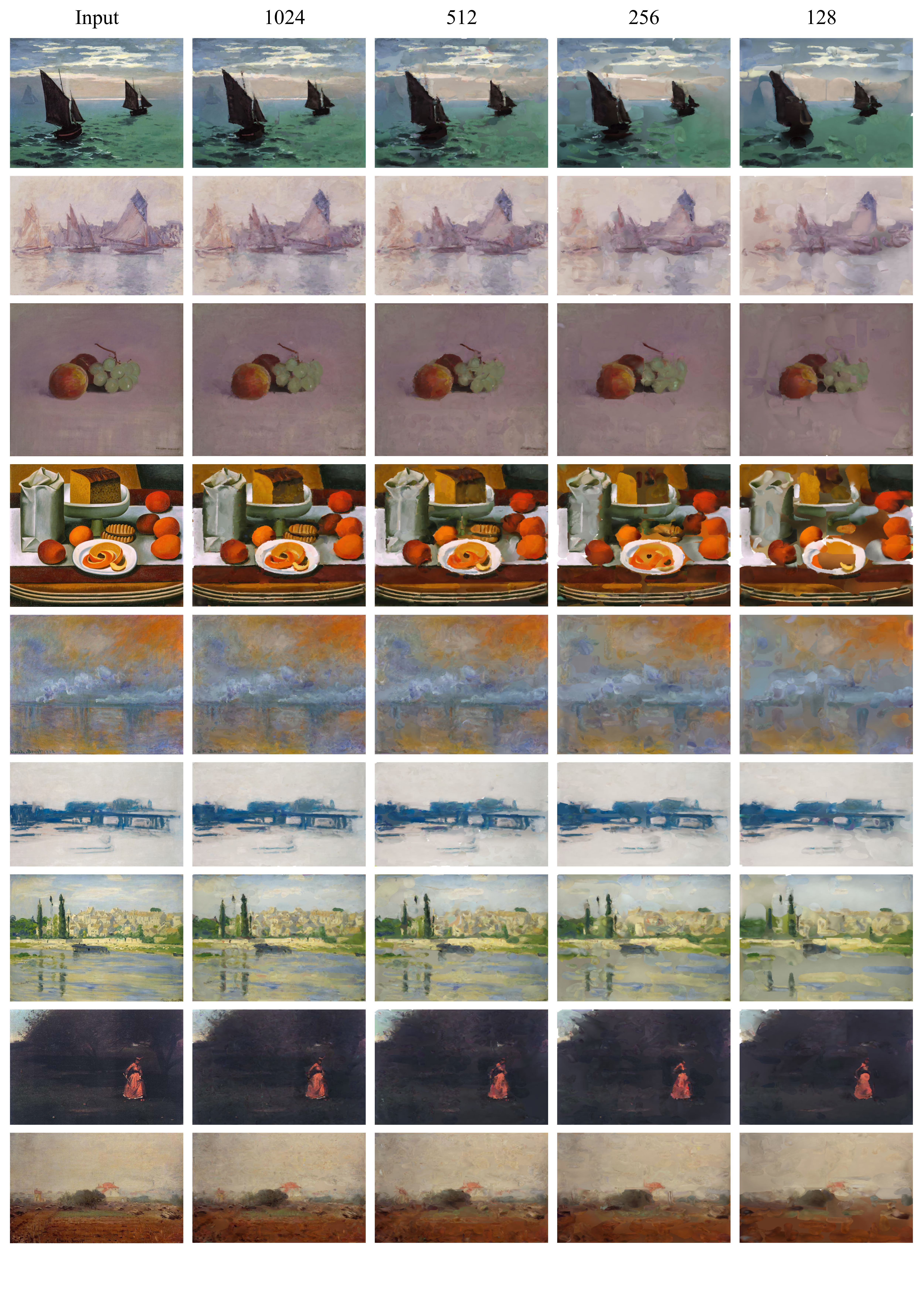}
\vspace{-35px}
\caption{More showcases from 128 to 1024 strokes in oil paintings with the proposed method.}
\label{fig:oil1}
\end{figure}

\begin{figure}
\centering
\includegraphics[width=0.95\textwidth]{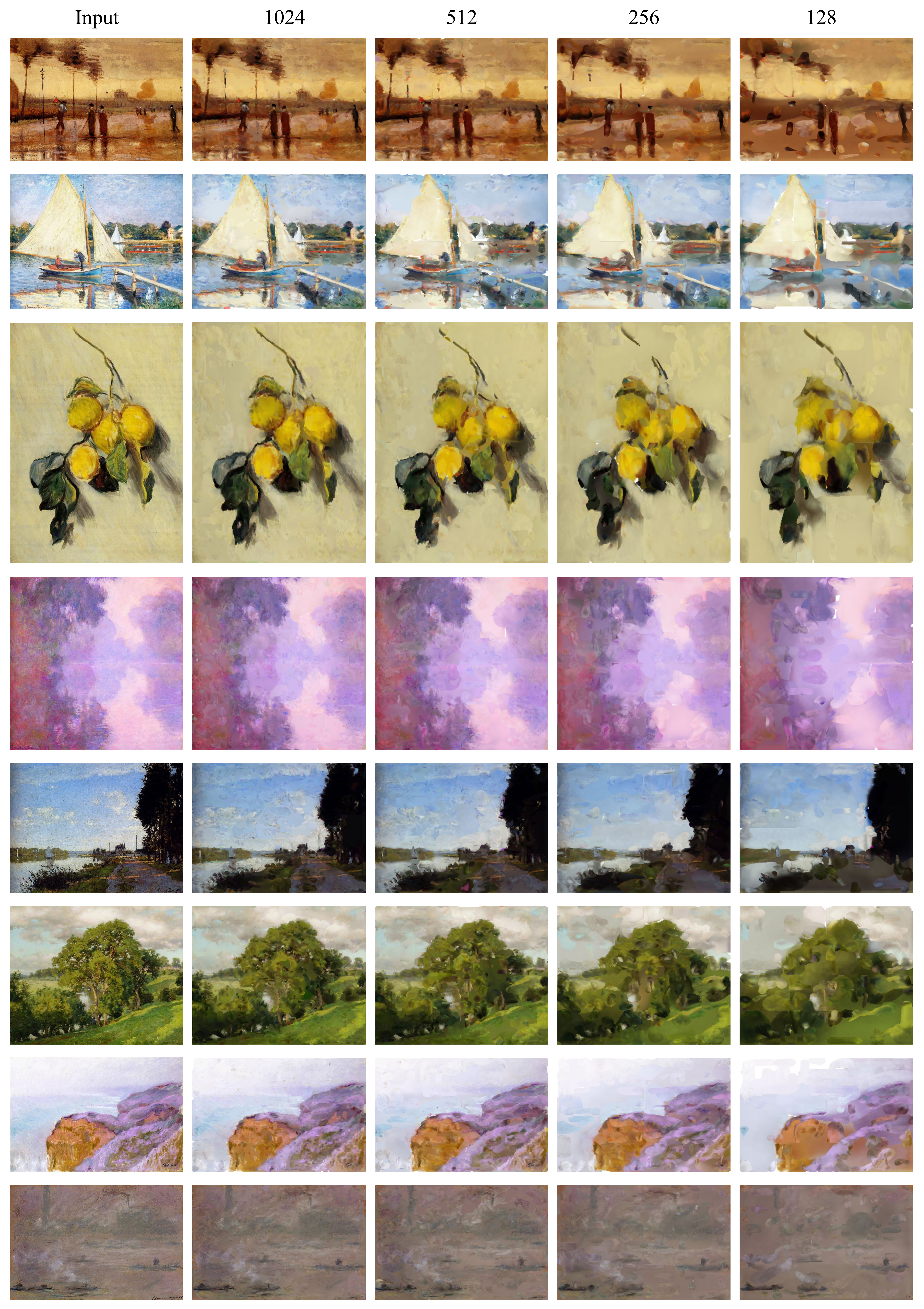}
\vspace{-10px}
\caption{More showcases from 128 to 1024 strokes in oil paintings with the proposed method.}
\label{fig:oil2}
\end{figure}

\begin{figure}
\centering
\includegraphics[width=0.95\textwidth]{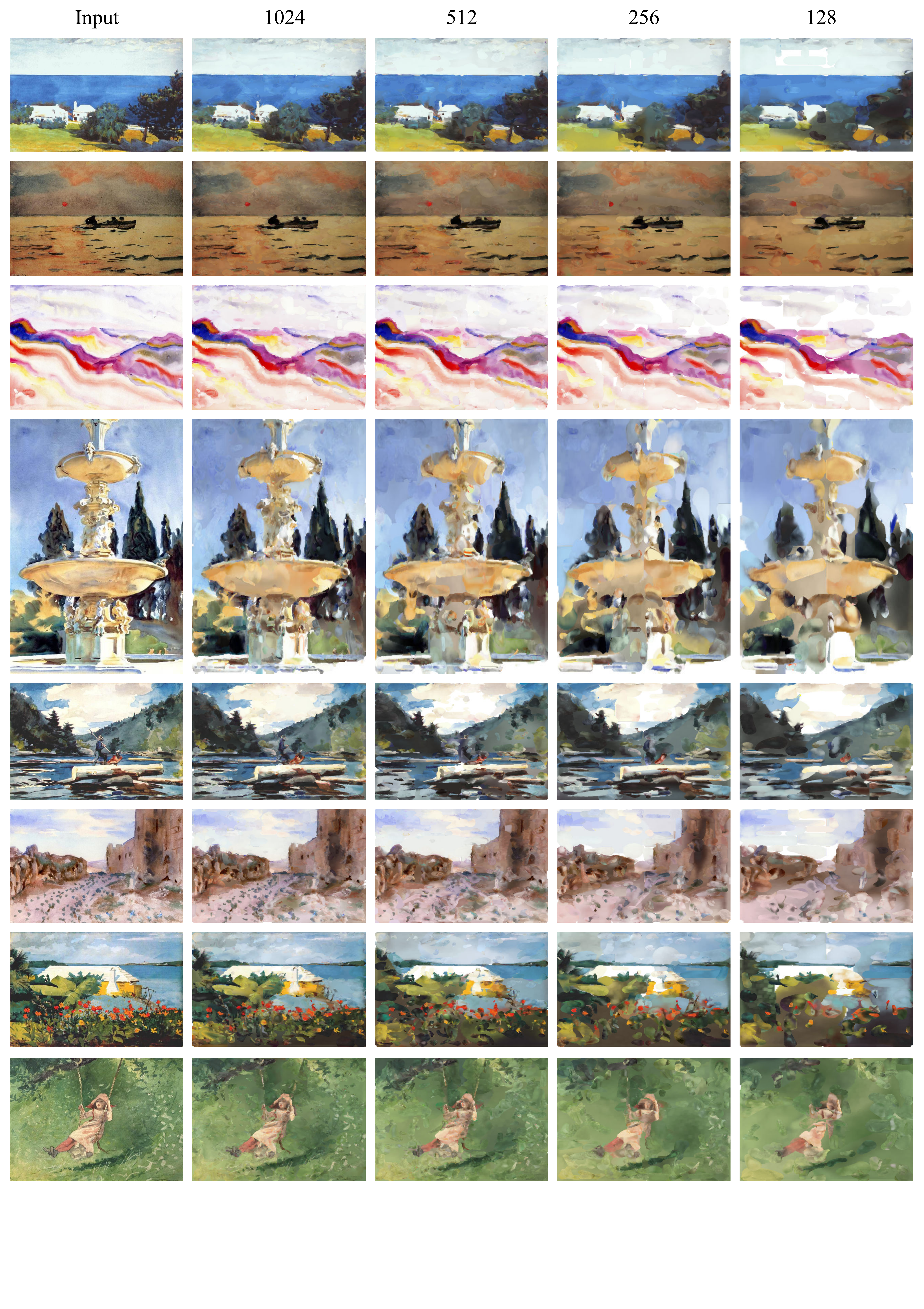}
\vspace{-65px}
\caption{More showcases from 128 to 1024 strokes in watercolor paintings with the proposed method.}
\label{fig:watercolor1}
\end{figure}

\begin{figure}
\centering
\includegraphics[width=0.95\textwidth]{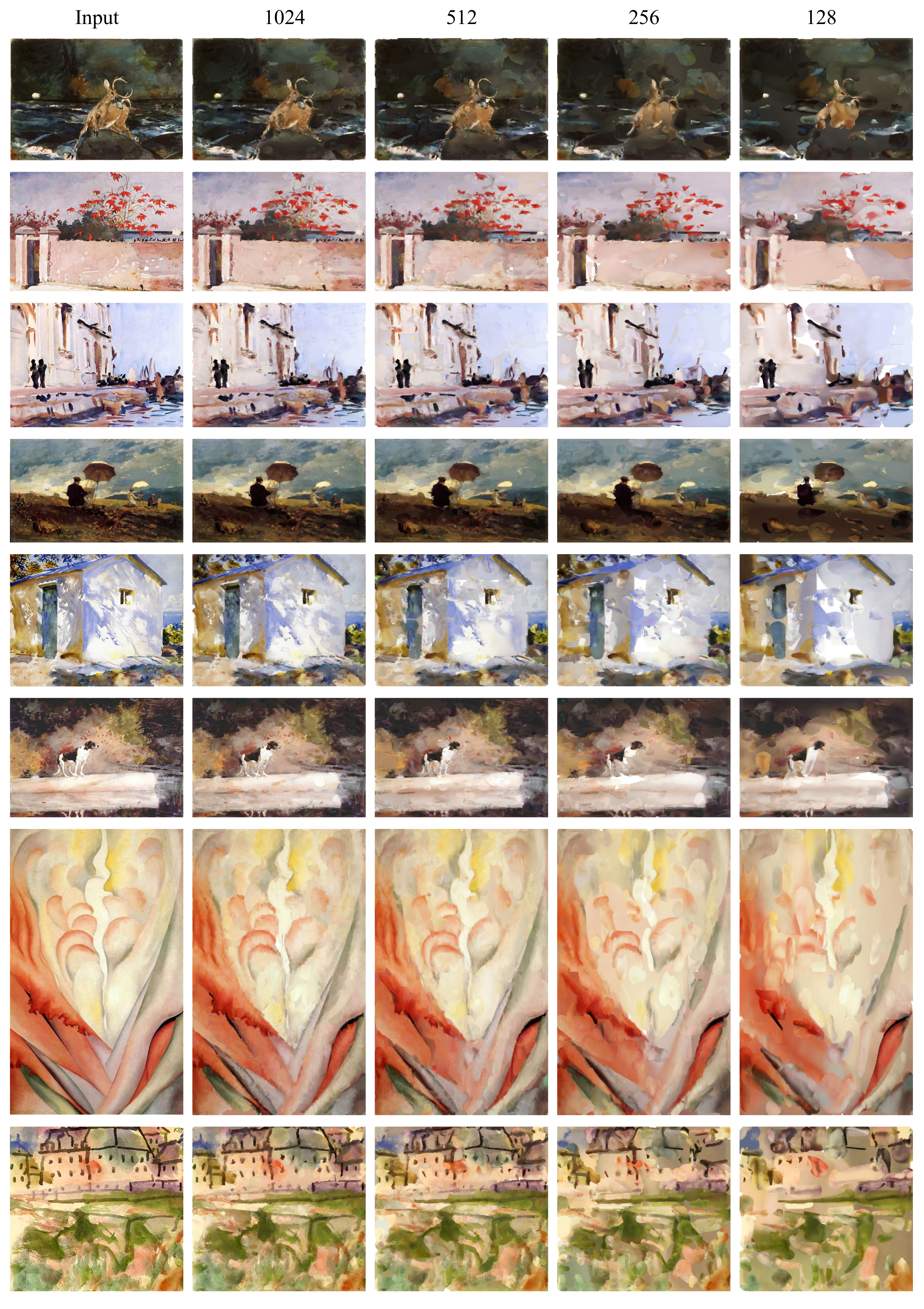}
\vspace{-15px}
\caption{More showcases from 128 to 1024 strokes in watercolor paintings with the proposed method.}
\label{fig:watercolor2}
\end{figure}

\section{More Qualitative Results}
\begin{figure}
\centering
\includegraphics[width=0.98\textwidth]{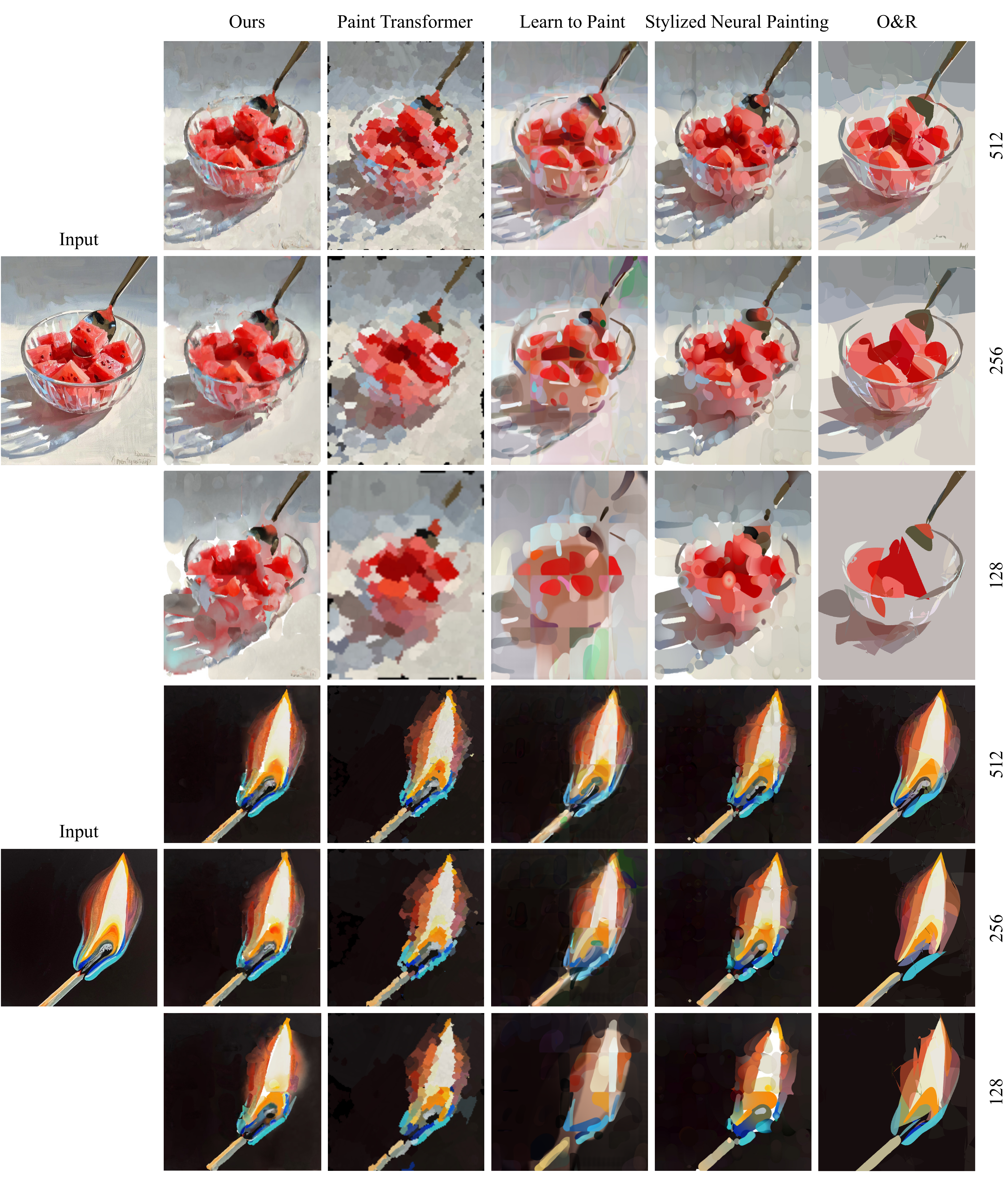}
\vspace{-10px}
\caption{More qualitative results from 128 to 512 strokes in digital paintings.}
\label{fig:qua_digital_2}
\end{figure}

\begin{figure}
\centering
\includegraphics[width=0.98\textwidth]{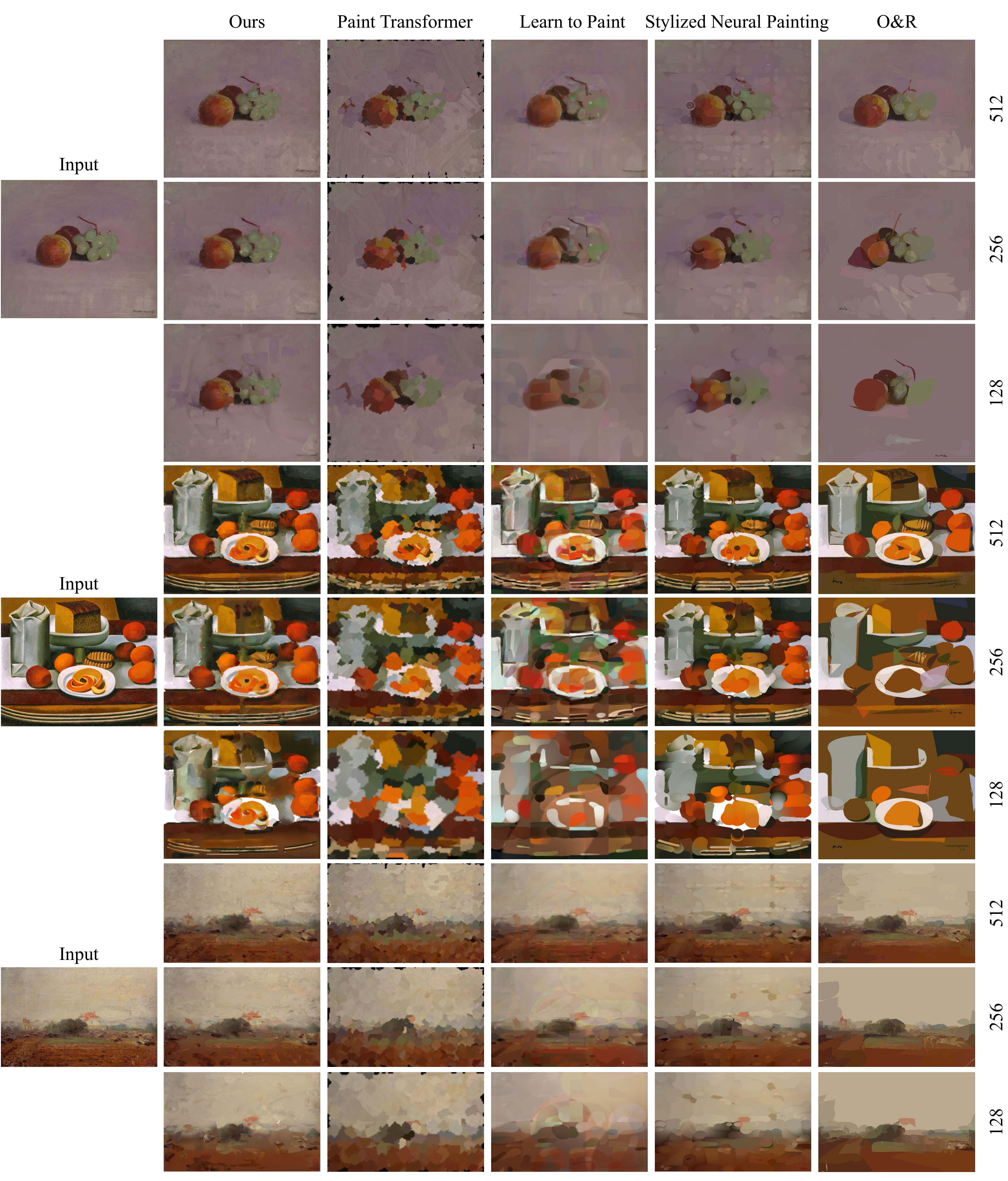}
\vspace{-10px}
\caption{More qualitative results from 128 to 512 strokes in oil paintings.}
\label{fig:qua_oil_1}
\end{figure}

\begin{figure}
\centering
\includegraphics[width=0.98\textwidth]{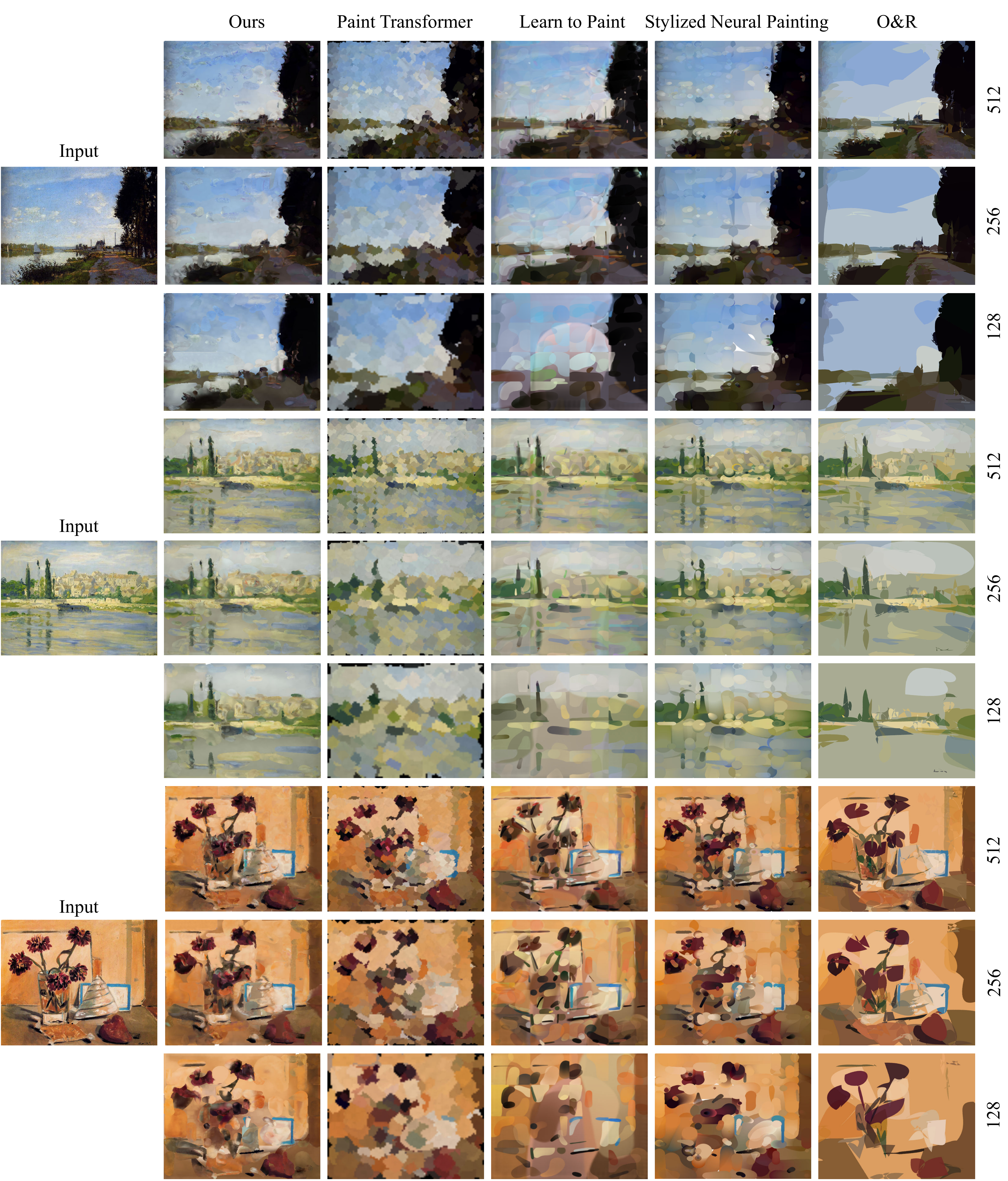}
\vspace{-5px}
\caption{More qualitative results from 128 to 512 strokes in oil paintings.}
\label{fig:qua_oil_2}
\end{figure}

\begin{figure}
\centering
\includegraphics[width=0.98\textwidth]{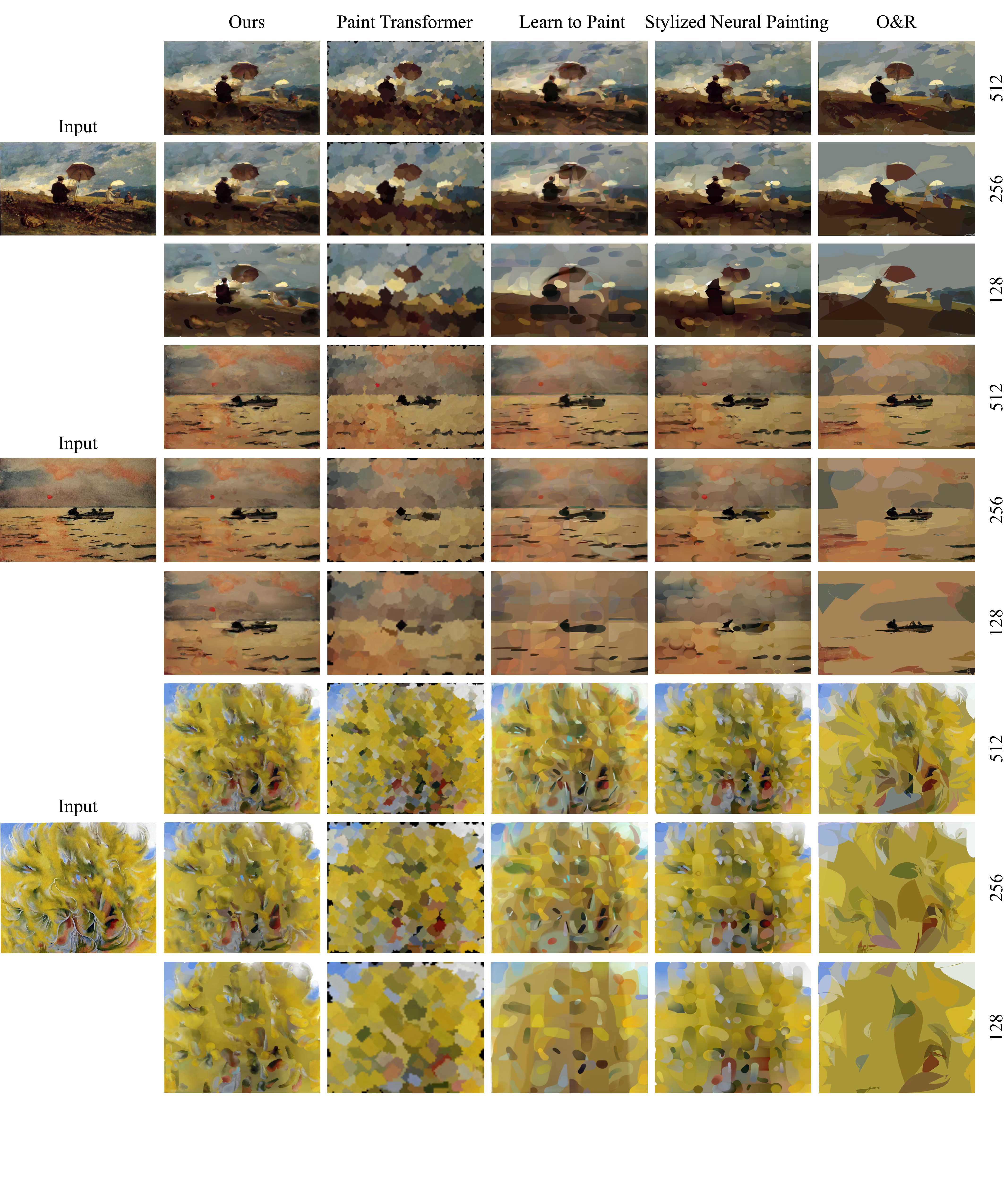}
\vspace{-45px}
\caption{More qualitative results from 128 to 512 strokes in watercolor paintings.}
\label{fig:qua_watercolor_1}
\end{figure}

\begin{figure}
\centering
\includegraphics[width=0.98\textwidth]{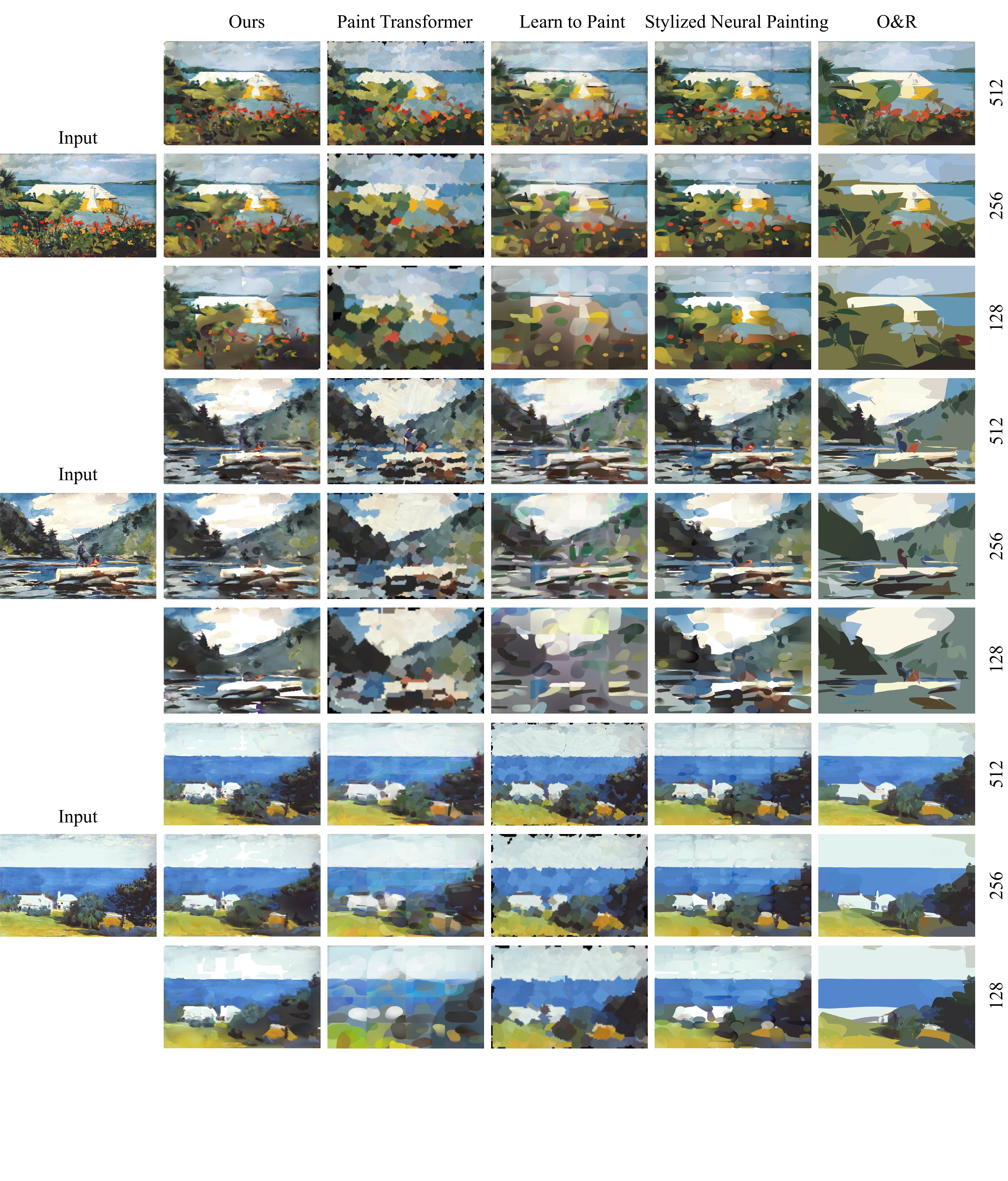}
\vspace{-60px}
\caption{More qualitative results from 128 to 512 strokes in watercolor paintings.}
\label{fig:qua_watercolor_2}
\end{figure}

\begin{figure}
\centering
\includegraphics[width=0.98\textwidth]{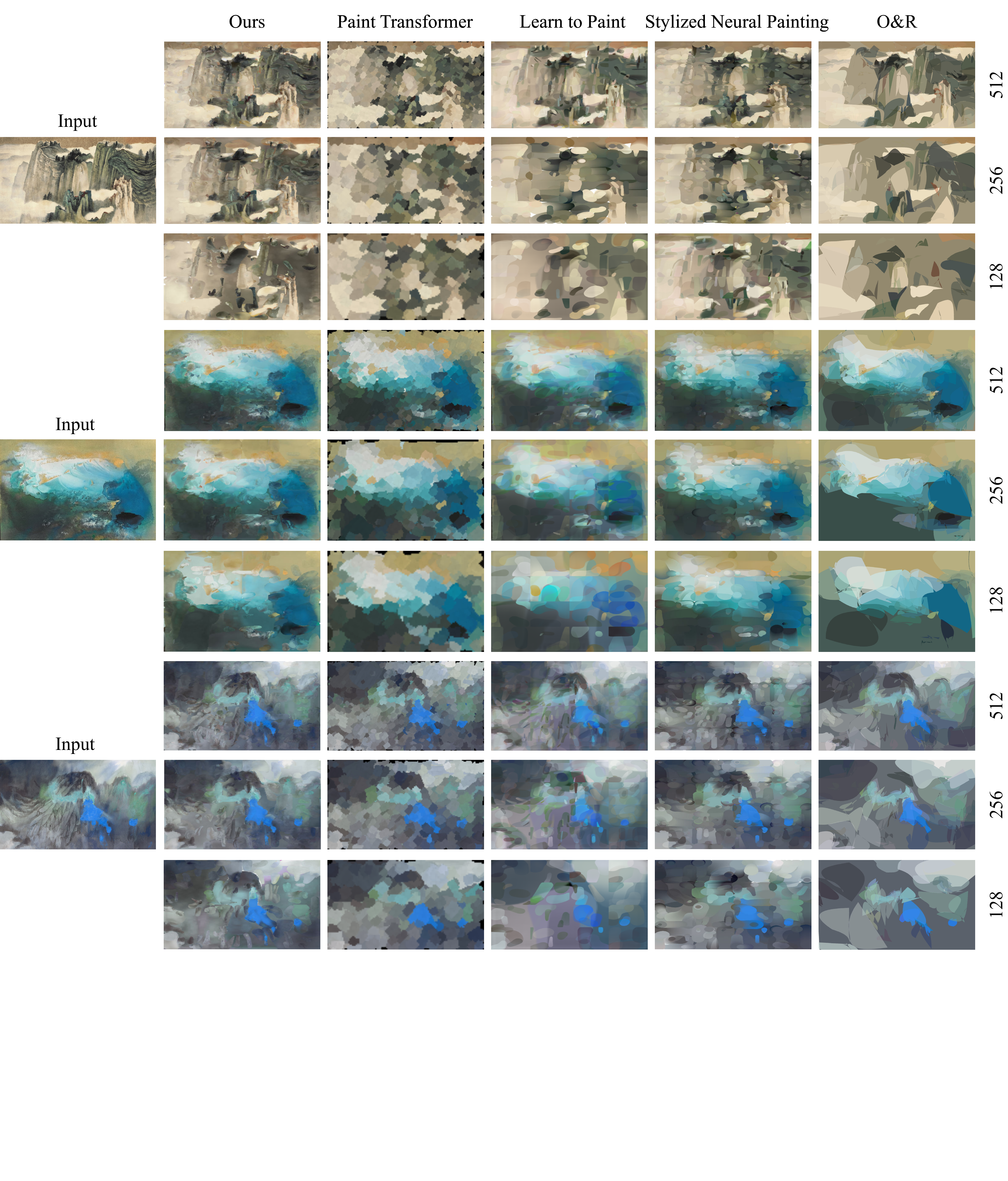}
\vspace{-105px}
\caption{More qualitative results from 128 to 512 strokes in ink paintings.}
\label{fig:qua_ink_1}
\end{figure}

\begin{figure}
\centering
\includegraphics[width=0.98\textwidth]{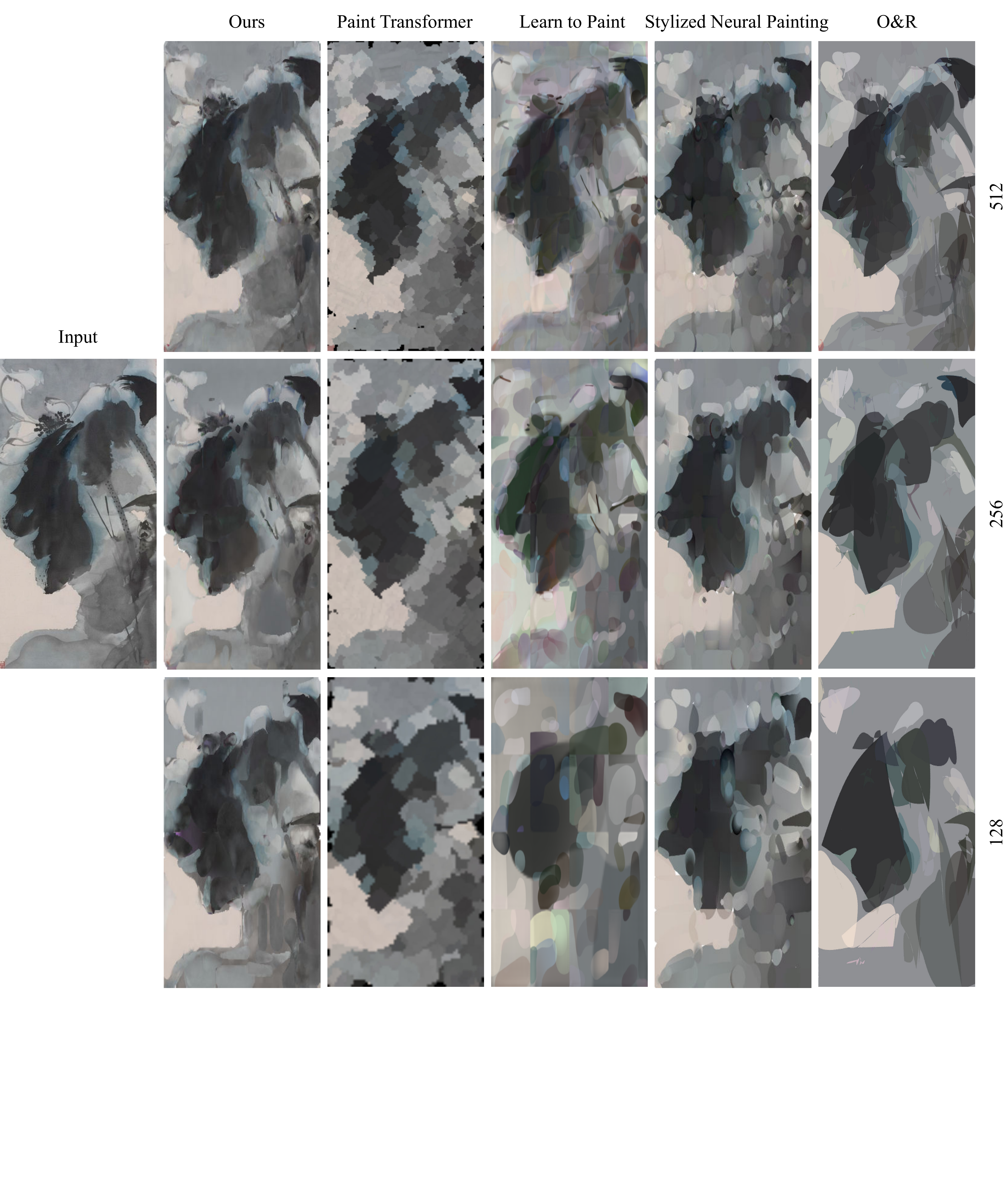}
\vspace{-90px}
\caption{More qualitative results from 128 to 512 strokes in ink paintings.}
\label{fig:qua_ink_2}
\end{figure}

We provide additional qualitative comparisons against Learning to Paint~\cite{huang2019learning}, PaintTransformer~\cite{liu2021paint}, Stylized Neural Painting~\cite{zou2021stylized}, and O\&R~\cite{hirschorn2024optimize}, illustrating reconstructions from 128 to 512 strokes across four representative painting styles: digital (Fig.~\ref{fig:qua_digital_2}), oil (Fig.~\ref{fig:qua_oil_1}, Fig.~\ref{fig:qua_oil_2}), watercolor (Fig.~\ref{fig:qua_watercolor_1}, Fig.~\ref{fig:qua_watercolor_2}), and ink (Fig.~\ref{fig:qua_ink_1}, Fig.~\ref{fig:qua_ink_2}).
Our method consistently achieves coherent structural alignment, smooth tonal transitions, and realistic painterly effects under varying stroke budgets, demonstrating its robustness and adaptability across different artistic styles.


\end{document}